# Minimax rates for cost-sensitive learning on manifolds with approximate nearest neighbours*

**Henry W J Reeve**† HENRY.REEVE@MANCHESTER.AC.UK
**Gavin Brown** GAVIN.BROWN@MANCHESTER.AC.UK
*School of Computer Science*
*University of Manchester*
*Manchester M13 9PL*

## Abstract

We study the approximate nearest neighbour method for cost-sensitive classification on low-dimensional manifolds embedded within a high-dimensional feature space. We determine the minimax learning rates for distributions on a smooth manifold, in a cost-sensitive setting. This generalises a classic result of Audibert and Tsybakov. Building upon recent work of Chaudhuri and Dasgupta we prove that these minimax rates are attained by the approximate nearest neighbour algorithm, where neighbours are computed in a randomly projected low-dimensional space. In addition, we give a bound on the number of dimensions required for the projection which depends solely upon the *reach* and dimension of the manifold, combined with the regularity of the marginal.

## 1. Introduction

The nearest neighbour method is a simple and intuitive approach to classification with numerous strong theoretical properties. A classical result of Stone (1977) gives convergence to the Bayes error. More recently, Chaudhuri and Dasgupta (2014) demonstrated that the nearest neighbour method adapts to the unknown level of noise, expressed as a margin condition. Indeed, under Tsybakov's margin condition, the risk converges at the minimax optimal rates for binary classification identified by Audibert et al. (2007), which require that the marginal is absolutely continuous with respect to the Lebesgue measure. In this work we move the analysis of minimax rates for classification closer to practical settings encountered in machine learning applications.

High dimensional feature spaces occur in many machine learning applications from computer vision through to genome analysis and natural language processing. Whilst the dimensionality of the feature space may be high, the data itself is often constrained to a low-dimensional manifold. This renders the assumption of an absolutely continuous marginal distribution inappropriate. As we shall see, optimal classification rates are dependent upon the intrinsic complexity of the manifold.

High-dimensional feature spaces also give rise to computational challenges. Indeed, an exact nearest neighbour search is often prohibitively expensive when the feature space is high dimensional

* The authors would like to thank Ata Kaban for numerous useful conversations which greatly improved the paper. We would also like to thank Joe Mellor and the anonymous reviewers for useful feedback. In addition we gratefully acknowledge the support of the EPSRC for the LAMBDA project (EP/N035127/1) and the Manchester Centre for Doctoral Training (EP/1038099/1).

† Corresponding author

(Indyk and Motwani (1998)). An efficient approach to dealing with these computational challenges is random projections (Dirksen (2016)). Kabán (2015) demonstrated that the randomly projected nearest neighbour method is capable of exploiting low intrinsic complexity within the data set in the distribution-free setting. However, these distribution-free bounds are non-optimal when Tsybakov's margin condition holds. We provide optimal distribution-dependent bounds for the approximate nearest neighbour method.

The seminal works of Audibert et al. (2007) and Chaudhuri and Dasgupta (2014) target the overall classification risk. In doing so they implicitly make the assumption of symmetric costs. However, many real-world applications, from medical diagnosis through to fraud detection, have asymmetric costs: Different mis-classification errors incur different costs (see Elkan (2001); Dmochowski et al. (2010)). The nearest neighbour method can be straight-forwardly adapted to the *cost-sensitive* setting in which asymmetric costs are taken into account (see Section 5). In this work we analyze the nearest neighbour method in high-dimensional cost-sensitive settings, in keeping with our goal of bringing the analysis of nearest neighbour methods closer to practical settings encountered in machine learning applications.

Building upon previous work of Audibert et al. (2007); Chaudhuri and Dasgupta (2014); Kabán (2015), we give optimal, distribution-dependent, and *cost-sensitive* bounds for the approximate nearest neighbour method, when the data is concentrated on a low-dimensional manifold. Specifically, we provide the following.

- We determine the minimax learning rates for a natural family of distributions supported on embedded manifolds, in a multi-class cost-sensitive setting (Section 4);
- We demonstrate that these rates are attained by an approximate nearest neighbour algorithm, where neighbours are computed in a low-dimensional randomly projected space (Section 5);
- We give a bound upon on the number of dimensions required for the projection to attain optimal learning rates which depends solely upon the reach and dimension of the manifold, combined with the regularity of the marginal distribution (Section 5).

We begin by introducing some notation in Sections 2 and 3 before stating our main results in Sections 4,5 and 6. Detailed proofs are provided in the Appendix: Sections A, B, C and D.

## 2. Background I: Classification with nearest neighbours

In this section we introduce our notation and give the relevant background on the nearest neighbour method and distribution dependent bounds. This will lead us into a discussion on high dimensional data and manifolds. We will bring these two strands together in Sections 4 and 5.

### 2.1. The classification problem

Suppose we have a distribution $\mathbb{P}$ over a $\mathcal{Z} = \mathcal{X} \times \mathcal{Y}$, where $(\mathcal{X}, \rho)$ is a metric space and $\mathcal{Y} = \{1, \cdots, L\}$ is a discrete space of labels. We let $\Delta(\mathcal{Y}) \subset \mathbb{R}^L$ denote the $(L-1)$-simplex consisting of probability vectors over $\mathcal{Y}$. The distribution $\mathbb{P}$ over $\mathcal{Z}$ is determined by a marginal distribution $\mu$ on $\mathcal{X}$, and a conditional probability specified by $\eta : \mathcal{X} \to \Delta(\mathcal{Y})$ where for each $x \in \mathcal{X}$, $y \in \mathcal{Y}$, $\eta(x)_y = \mathbb{P}[Y = y|X = x]$. We let $\mathbb{E}$ denote expectation over random pairs $(X, Y) \sim \mathbb{P}$. We take a fixed cost matrix $\Phi$ with entries $\phi_{ij} \geq 0$ denoting the cost incurred by predicting class $i$ when the true label is class $j$. Following Elkan (2001) we shall say that a cost-matrix $\Phi$ is *reasonable* when $\phi_{ii} < \phi_{ji}$ for all $i, j$ with $j \neq i$. Often it is assumed that all mistakes are equally expensive and the cost matrix $\Phi_{0,1}$ with all diagonal entries equal to 0 and all non-diagonal entries equal to 1 is used. However, there are many application domains where the assumption of a symmetric cost matrix is highly inaccurate (Dmochowski et al. (2010)). The class of *reasonable* cost matrices provides a more general assumption applicable to a wide range of cost-sensitive scenarios. In particular, the class of reasonable cost matrices generalises the class considered by Zhang (2004).

Given a cost matrix $\Phi$, the risk $R(f)$ of a classifier $f : \mathcal{X} \to \mathcal{Y}$ is defined by $R(f) = \mathbb{E}\left[\phi_{f(X),Y}\right]$. The Bayes risk $R^*$ is defined by $R^* = \inf\{R(f) : f : \mathcal{X} \to \mathcal{Y} \text{ is Borel}\}$. Our goal is to obtain a classifier $f : \mathcal{X} \to \mathcal{Y}$ with $R(f)$ as close as possible to $R^*$. Whilst we do not have direct access to the distribution $\mathbb{P}$, we do have access to a random data set $\mathcal{D}_n = \{Z_1, \cdots, Z_n\}$ with $Z_i$ selected independently according to $\mathbb{P}$. Equivalently, $\mathcal{D}_n \sim \mathbb{P}^n$ where $\mathbb{P}^n$ denotes the product measure $\prod_{i=1}^n \mathbb{P}$ on $\mathcal{Z}^n$. We let $\mathbb{E}_n$ denote expectation over data sets $\mathcal{D}_n \sim \mathbb{P}^n$ and let $\mathcal{F}_n$ denote the set of feature vectors in the training data ie. $\mathcal{F}_n = \{X_1, \cdots, X_n\}$ where $\mathcal{D}_n = \{(X_1, Y_1), \cdots, (X_n, Y_n)\}$.

### 2.2. Distribution dependent bounds for classification

One of the central goals of statistical learning theory is to establish optimal bounds on the risk of a classifier, under natural assumptions on the distribution. The seminal work of Audibert et al. (2007) established optimal bounds for a class of distributions supported on regular sets of positive Lebesgue measure, satisfying a margin condition. To recall these results we require some notation.

**Definition 2.1 (Regular sets and measures)** *Suppose we have a measure $\upsilon$ on the metric space $(\mathcal{X}, \rho)$. A subset $A \subset \mathcal{X}$ is said to be a $(c_0, r_0)$-regular set with respect to the measure $\upsilon$ if for all $x \in A$ and all $r \in (0, r_0)$ we have $\upsilon(A \cap B_r(x)) \geq c_0 \cdot \upsilon(B_r(x))$, where $B_r(x)$ denotes the open metric ball of radius $r$, centred at $x$. A measure $\mu$ with support supp$(\mu) \subset \mathcal{X}$ is said to be $(c_0, r_0, \nu_{\min}, \nu_{\max})$-regular measure with respect to $\upsilon$ if supp$(\mu)$ is a $(c_0, r_0)$-regular set with respect to $\upsilon$ and $\mu$ is absolutely continuous with respect to $\upsilon$ with Radon-Nikodym derivative $\nu(x) = d\mu(x)/d\upsilon(x)$, such that for all $x \in$ supp$(\mu)$ we have $\nu_{\min} \leq \nu(x) \leq \nu_{\max}$.*

The assumption of a regular marginal ensures that with high probability there are a large number of training points in the vicinity of the average test point.

**Definition 2.2 (Hölder conditional)** *The conditional $\eta$ is said to be* Hölder *continuous with constants $(\alpha, C_\alpha)$ if for $\mu$ almost every $x_0, x_1 \in \mathcal{X}$ we have $\|\eta(x_0) - \eta(x_1)\|_\infty \leq C_\alpha \cdot \rho(x_0, x_1)^\alpha$.*

The assumption of a Hölder conditional ensures that the proximity between a test point and its closest training points is reflected in a similar value for the conditional probability. Hence, higher Hölder exponents correspond to faster learning rates. Audibert and Tsybakov also considered higher order smoothness conditions and showed that in such settings even faster learning rates are attainable (see (Audibert et al., 2007, Section 2 & Section 3, Theorem 3.3)). However, in this work we restrict our attention to the Hölder condition given in Definition 2.2.

**Definition 2.3 (Tsybakov's Margin condition)** *Suppose that $\mathcal{Y} = \{1, 2\}$ and $\Phi = \Phi_{0,1}$. We shall say that $\mathbb{P}$ satisfies the margin condition with constants $(C_\beta, \beta)$ if for all $\zeta > 0$ we have $\mu\left(\left\{x \in \mathcal{X} : 0 < \left|\eta(x)_1 - \frac{1}{2}\right| \leq \zeta\right\}\right) \leq C_\beta \cdot \zeta^\beta$.*

The margin condition bounds the probability that the labels of training points in the vicinity of the test point will disagree with the mode of the conditional label distribution. We generalise the condition to an arbitrary $L \times L$ cost matrix as follows. Given $\boldsymbol{n} \in \Delta(\mathcal{Y})$ we let $\mathcal{Y}^*_\Phi(\boldsymbol{n}) \subset \mathcal{Y}$ denote the set of labels with minimal associated cost. That is, $\mathcal{Y}^*_\Phi(\boldsymbol{n}) = \text{argmin}_{y\in\mathcal{Y}}\left\{e(y)^T\Phi\,\boldsymbol{n}\right\}$, where $e(y) \in \{0,1\}^{L\times 1}$ denotes the 'one-hot-encoding', satisfying $e(y)_y = 1$ and $e(y)_l = 0$ for $l \neq y$. Given $\boldsymbol{n} \in \Delta(\mathcal{Y})$ we define

$$M_\Phi(\boldsymbol{n}) = \min\left\{(e(y_1) - e(y_0))^T\,\Phi\,\boldsymbol{n} : y_0 \in \mathcal{Y}^*_\Phi(\boldsymbol{n}),\ y_1 \in \mathcal{Y}\backslash\mathcal{Y}^*_\Phi(\boldsymbol{n})\right\}.$$

**Definition 2.4 (Margin condition)** *We shall say that $\mathbb{P}$ satisfies the margin condition with constants $(C_\beta, \beta, \zeta_{\max})$ if for all $\zeta \in (0, \zeta_{\max})$ we have $\mu(\{x \in \mathcal{X} : M_\Phi(\eta(x)) \leq \zeta\}) \leq C_\beta \cdot \zeta^\beta$.*

The Definitions 2.1, 2.2, 2.4 characterise a natural family of distributions as follows.

**Definition 2.5 (Measure classes)** *Fix positive constants $\alpha, \beta, r_0, c_0, \nu_{\min}, \nu_{\max}, \zeta_{\max}, C_\alpha, C_\beta$ and let $\Gamma = \langle(c_0, r_0, \nu_{\min}, \nu_{\max}), (\beta, C_\beta, \zeta_{\max}), (\alpha, C_\alpha)\rangle$. We let $\mathcal{P}_\Phi(\upsilon, \Gamma)$ denote the class of all probability measures $\mathbb{P}$ on $\mathcal{Z} = \mathcal{X} \times \mathcal{Y}$ such that*

- *$\mu$ is a $(c_0, r_0, \nu_{\min}, \nu_{\max})$-regular measure with respect to $\upsilon$,*
- *$\eta$ is Hölder continuous with constants $(\alpha, C_\alpha)$,*
- *$\mathbb{P}$ satisfies the margin condition with constants $(\beta, C_\beta, \zeta_{\max})$.*

Audibert et al. (2007) gave the following minimax result for classes of the form $\mathcal{P}_{\Phi_{0,1}}(\mathcal{L}_d, \Gamma)$, where $\mathcal{L}_d$ denotes the $d$-dimensional Lebesgue measure on $\mathbb{R}^d$. The Euclidean metric $\rho$ on $\mathbb{R}^d$ is given by $\rho(x_0, x_1) = \|x_0 - x_1\|_2$, where $\|\cdot\|_2$ denotes the Euclidean norm. To state the result we must distinguish between the estimation procedure $\hat{f}^{\text{est}}_n \in \left(\mathcal{Y}^\mathcal{X}\right)^{\mathcal{Z}^n}$ and the classifier $\hat{f}_n \in \mathcal{Y}^\mathcal{X}$. The estimation procedure $\hat{f}^{\text{est}}_n : \mathcal{Z}^n \to \mathcal{Y}^\mathcal{X}$ is a mapping which takes a data set $\mathcal{D}_n \in \mathcal{Z}^n$ and outputs a classifier $\hat{f}_n : \mathcal{X} \to \mathcal{Y}$, which implicitly depends upon the particular data set $\mathcal{D}_n$.

**Theorem 1 (Audibert et al. (2007))** *Take $d \in \mathbb{N}$, $\mathcal{X} = \mathbb{R}^d$, let $\rho$ denote the Euclidean metric and set $\mathcal{Y} = \{1, 2\}$. There exists positive constants $C_0, R_0, V_-, V_+$ such that for all $c_0 \in (0, C_0)$, $r_0 \in (0, R_0)$, $\nu_{\min} \in (0, V_-)$, $\nu_{\max} \in (V_+, \infty)$, $\alpha \in (0, 1)$, $\beta \in (0, d/\alpha)$, $C_\alpha, C_\beta > 0$, $\zeta_{\max} > 0$, if we take $\Gamma = \langle(c_0, r_0, \nu_{\min}, \nu_{\max}), (\beta, C_\beta, \zeta_{\max}), (\alpha, C_\alpha)\rangle$, we have*

$$\inf\left\{\sup\left\{\mathbb{E}_{\mathbb{P}_n}\left[R\left(\hat{f}_n\right)\right] - R^* : \mathbb{P} \in \mathcal{P}_{\Phi_{0,1}}(\mathcal{L}_d, \Gamma)\right\} : \hat{f}^{est}_n \in \left(\mathcal{Y}^\mathcal{X}\right)^{\mathcal{Z}^n}\right\} = \Theta\left(n^{-\frac{\alpha(1+\beta)}{2\alpha+d}}\right),$$

*with upper and lower constants determined solely by $d$ and $\Gamma$.*

Here we use standard complexity notation (Cormen, 2009, Chapter 3). In the proof of Theorem 1 Audibert et al. also identified a classifier based on kernel density estimation which attains the minimax optimal convergence rates Audibert et al. (2007). This raises the interesting question of which other classifiers attain these rates.

### 2.3. The nearest neighbour classifier

The nearest neighbour classifier is constructed as follows. Given $k \leq n$ we let $S_k^\circ(x, \mathcal{F}_n) \subseteq \{1, \cdots, n\}$ denote a set of for $k$-nearest neighbour indices. That is, $\mathcal{I} = S_k^\circ(x, \mathcal{F}_n)$ minimises $\max\{\rho(x, X_i) : i \in \mathcal{I}\}$ over all sets $\mathcal{I} \subseteq \{1, \cdots, n\}$ with $\#\mathcal{I} = k$. The $k$-nearest classifier is defined by $\hat{f}_{k,n}(x) = \text{Mode}(\{Y_i : i \in S_k^\circ(x, \mathcal{F}_n)\})$.

Despite its simplicity the nearest neighbour classifier has strong theoretical properties. Shalev-Shwartz and Ben-David gave an elegant proof that the generalisation error of the 1-nearest neighbour classifier converges to at most $2 \cdot R^* + O(n^{-1/(1+d)})$, for Lipchitz $\eta$, without any assumptions on the marginal $\mu$ (Shalev-Shwartz and Ben-David, 2014, Chapter 19). This approach can be extended to all metric spaces of doubling dimension $d$ (see Kontorovich and Weiss (2014)). Chaudhuri and Dasgupta gave distribution dependent bounds for the nearest neighbour method on general metric spaces (Chaudhuri and Dasgupta (2014)). Rather than relying directly upon the Hölder continuity of the conditional, Chaudhuri and Dasgupta introduced the following smoothness condition, which is especially suited to non-parametric classification. Given $x \in \mathcal{X}$ and $r > 0$ we let $B_r(x) = \{\tilde{x} \in \mathcal{X} : \rho(x, \tilde{x}) < r\}$, and $\overline{B_r(x)} = \{\tilde{x} \in \mathcal{X} : \rho(x, \tilde{x}) \leq r\}$.

**Definition 2.6 (Measure-smooth conditional)** *The conditional $\eta$ is said to be measure-smooth with constants $(\lambda, C_\lambda)$ if for $\mu$ almost every $x_0, x_1 \in \mathcal{X}$ we have*

$$\|\eta(x_0) - \eta(x_1)\|_\infty \leq C_\lambda \cdot \mu\left(B_{\rho(x_0,x_1)}(x_0)\right)^\lambda .$$

Chaudhuri and Dasgupta (2014) proved that whenever the conditional is measure-smooth and the Tsybakov margin condition holds, then for $\Phi = \Phi_{0,1}$ and $\mathcal{Y} = \{1, 2\}$, the risk of the nearest neighbour method converges to the Bayes error at a rate $O\left(n^{-\lambda(1+\beta)/(2\lambda+1)}\right)$. It follows that the nearest neighbour classifier attains the optimal convergence rates for $\mathbb{P} \in \mathcal{P}_{\Phi_{0,1}}\left(\mathcal{L}^d, \Gamma\right)$ given in Theorem 1 (Chaudhuri and Dasgupta, 2014, Lemma 2).

## 3. Background II: High dimensional data

A wide variety of machine learning application domains from computer vision through to genome analysis involve extremely high-dimensional feature spaces. Nonetheless, statistical regularities in the data often mean that its intrinsic complexity is often much lower than the number of features. A natural approach to modelling this low intrinsic complexity is to assume that the data lies on a manifold (see Roweis and Saul (2000); Tenenbaum et al. (2000); Park et al. (2015)).

### 3.1. Manifolds

We shall consider a compact $C^\infty$-smooth sub-manifold of $\mathcal{M} \subset \mathbb{R}^d$ of dimension $\gamma$ (see Lee (1997)). The manifold is endowed with two natural metrics. Given a pair of points $x_0, x_1 \in \mathcal{M}$, distances may be computed either with respect to the Euclidean metric $\rho(x_0, x_1) = \|x_0 - x_1\|_2$ (since $\mathcal{M} \subset \mathbb{R}^d$), or with respect to the geodesic distance induced by the manifold,

$$\rho_g(x_0, x_1) := \inf\left\{\int_0^1 \|c'(t)\|_2 : \quad c : [0, 1] \to \mathcal{M} \text{ is piecewise } C^1 \text{ with } c(0) = x_0 \ \& \ c(1) = x_1\right\}.$$

We shall make use of the concept of reach $\tau$ introduced by Federer (1959) and investigated by Niyogi et al. (2008). The reach $\tau$ of a manifold $\mathcal{M}$ is defined by

$$\tau := \sup\left\{r > 0 : \forall z \in \mathbb{R}^d \inf_{q\in\mathcal{M}}\{\|z-q\|_2\} < r \implies \exists!\, p \in \mathcal{M},\ \ \|z-p\|_2 = \inf_{q\in\mathcal{M}}\{\|z-q\|_2\}\right\}.$$

Note that Niyogi et al. (2008) refers the condition number $1/\tau$, which is the reciprocal of the reach $\tau$. We let $dV_{\mathcal{M}}$ denote the Riemannian volume element and $V_{\mathcal{M}}$ the Riemannian volume.

### 3.2. Minimax rates for nearest neighbours on manifolds

The study of minimax rates for data lying on a low-dimensional manifold has received substantial attention in the regression setting. In particular Kpotufe (2011) determined the minimax optimal rates for regression on metric spaces of a given intrinsic dimension and showed that $k$-nearest neighbour regression attains these rates. Regression with nearest neighbours has also been studied in the semi-supervised domain (Goldberg et al. (2009); Moscovich et al. (2017)). By combining Theorem 4 from Chaudhuri and Dasgupta (2014) with (Eftekhari and Wakin, 2015, Lemma 12) one obtains an upper bound on the risk for the binary k-nearest neighbour classifier on a manifold. However, Chaudhuri and Dasgupta (2014) do not present minimax lower bounds for classification on non-Euclidean spaces. Proving lower bounds on non-Euclidean spaces for classification is complicated by the necessity of constructing distributions which simultaneously satisfy the margin condition and have a marginal with a regular support.

### 3.3. Approximate nearest neighbours and random projections

High dimensional data is computationally problematic for the nearest neighbour method. The naive nearest neighbour search depends linearly on the number of dimensions. More sophisticated solutions which are logarithmic in the number of examples lead to either a time or space complexity which is exponential in the number of features (Andoni and Indyk (2006)). Thus, nearest neighbour classification based on an exact nearest neighbour search is computationally prohibitive for high dimensional data sets with many examples. A popular and computationally tractable alternative is to use approximate nearest neighbours (Indyk and Motwani (1998)).

Given $\theta \geq 1$, a family of mappings $S = \{S_k\}_{k\in\mathbb{N}}$ is said to generate $\theta$ approximate nearest neighbours if for each $x \in \mathcal{X}$, $n \in \mathbb{N}$ and $k \leq n$, $S_k(x, \mathcal{F}_n)$ is a subset of $\{1, \cdots, n\}$ with cardinality $k$ such that $\max\{\rho(x, X_i) : i \in S_k(x, \mathcal{F}_n)\} \leq \theta \cdot \max\{\rho(x, X_i) : i \in S_k^\circ(x, \mathcal{F}_n)\}$. The associated approximate nearest neighbour classifier is given by $\hat{f}_{k,n}^S(x) = \text{Mode}\left(\{Y_i : i \in S_k(x, \mathcal{F}_n)\}\right)$.

A highly efficient approach to generating approximate nearest neighbours is to apply a subgaussian random projection (see Dirksen (2016)). Given a random variable $u$, the subgaussian norm is given by $\|u\|_{\psi_2} := \inf\left\{\psi > 0 : \mathbb{E}_u\left[\exp\left(\|u\|_2^2/\psi^2\right)\right] \leq 2\right\}$. Whenever $\|u\|_{\psi_2} < \infty$ the random vector $u$ is said to be subgaussian. More generally, a $d$-dimensional random vector $\boldsymbol{v}$ is said to be subgaussian if $\|\boldsymbol{v}\|_{\psi_2} := \sup\left\{\|\boldsymbol{v}^T w\|_{\psi_2} : w \in \mathbb{R}^d\ \ \|w\|_2 \leq 1\right\} < \infty$. A random vector $\boldsymbol{v}$ is said to be isotropic if for all $w \in \mathbb{R}^d$ we have $\mathbb{E}_{\boldsymbol{v}}\left[\left(\boldsymbol{v}^T w\right)^2\right] = \|w\|_2^2$. By a subgaussian random projection $\varphi : \mathbb{R}^d \to \mathbb{R}^h$ we shall mean a random mapping constructed by taking $h$ independent and identically distributed subgaussian and isotropic random vectors $\boldsymbol{v}_1, \cdots, \boldsymbol{v}_h$, taking a random matrix $\boldsymbol{V} := [\boldsymbol{v}_1, \cdots, \boldsymbol{v}_h]^T$ and letting $\varphi(x) = \sqrt{(1/h)} \cdot \boldsymbol{V}\,x$. The subgaussian norm is extended to subgaussian random projections $\varphi$ by defining $\|\varphi\|_{\psi_2} := \|\boldsymbol{v}_1\|_{\psi_2}$.

**Example:** An interesting family of a subgaussian random projections is the set of 'data-base friendly' projections introduced by Achlioptas (2003). The entries of the matrix are chosen i.i.d from the set $\{-\sqrt{3}, 0, +\sqrt{3}\}$, with respective probabilities $1/6, 2/3, 1/6$. The high (expected) level of sparsity gives rise to projections which are efficient both to store and to apply.

Given a subgaussian random projection $\varphi$, we define an associated family of mappings $S(\varphi) = \{S_k^{\varphi}\}_{k\in\mathbb{N}}$ by letting $S_k^{\varphi}(x, \mathcal{F}_n)$ denote the set of $k$ nearest neighbours of $\varphi(x)$ in the randomly projected feature space $\mathbb{R}^h$, i.e. $S_k^{\varphi}(x, \mathcal{F}_n) = S_k^{\circ}(\varphi(x), \varphi(\mathcal{F}_n))$ where $\varphi(\mathcal{F}_n) = \{\varphi(X_1), \cdots, \varphi(X_n)\}$. When $S(\varphi)$ generates approximate nearest neighbours we let $\hat{f}_{k,n}^{\varphi}$ denote $\hat{f}_{k,n}^{S(\varphi)}$. The celebrated Johnson-Lindenstrauss theorem states that, given a finite data set $A \subset \mathbb{R}^d$, for $h = \Omega(\log(\#A))$, with high probability, a subgaussian random projection $\varphi : \mathbb{R}^d \to \mathbb{R}^h$ will be bi-Lipchitz on $A$ (Johnson and Lindenstrauss (1984); Matoušek (2008)). Moreover, the bi-Lipchitz property implies that the associated family of mappings $S(\varphi)$ generates approximate nearest neighbours (see Lemma D.4), a fact that was utilised by Indyk and Motwani (1998). Klartag and Mendelson (2005) showed that it is sufficient to take $h = \Omega\left(\gamma_{\text{tal}}^2(A)\right)$ where $\gamma_{\text{tal}}(A)$ quantifies the metric-complexity of $A$ (see Section G for details). Recently, Dirksen (2016) has built upon the work Klartag and Mendelsen to give a unified theory of dimensionality reduction, which will be critical for our main results. The results of Klartag and Mendelson (2005) and Dirksen (2016) are highly significant from a statistical learning theory perspective, since for many natural examples, such as when $\mathcal{X} \subset \mathbb{R}^d$ lies within a smooth manifold, the metric complexity $\gamma_{\text{tal}}(\mathcal{X})$ may be bounded independently of the cardinality of $\mathcal{X}$. In such cases we may bound the required number of projection dimensions $h$ independently of the number of examples $n$. Indeed, Kabán (2015) applied the results of Klartag and Mendelson (2005) to obtain improved bounds on the generalisation error of the approximate nearest neighbour classifier $\hat{f}_{k,n}^{\varphi}$ when $\mathcal{X} \subset \mathbb{R}^d$ and the metric complexity $\gamma_{\text{tal}}(\mathcal{X}) \ll d$. Kabán (2015) showed that given a sub-Gaussian random projection $\varphi : \mathbb{R}^d \to \mathbb{R}^h$ with $h = \Omega\left(\gamma_{\text{tal}}^2(\mathcal{X})\right)$, with high probability the generalisation error of the approximate 1-nearest neighbour classifier $\hat{f}_{1,n}^{\varphi}$ is bounded above by $2 \cdot R^* + O(n^{-1/(1+h)})$. Kabán (2015) also gives the same bound for the exact nearest neighbour classifier, dependent upon $\gamma_{\text{tal}}^2(\mathcal{X})$. Hence, convergence rates for both the approximate nearest neighbour classifier and the exact nearest neighbour classifier may be improved under the assumption of low-metric complexity in the distribution free setting.

### 3.4. Motivating questions

This raises the following questions. If we combine an assumption of low-intrinsic complexity with regularity assumptions analogous to those in Theorem 1, then what are the best possible rates? How do these rates depend upon geometric properties of the manifold? Are these rates the same for all *reasonable* cost matrices? Which algorithms attain these bounds?

## 4. Minimax rates for cost-sensitive learning on manifolds

In this section we shall give minimax learning rates for cost-sensitive learning on manifolds. Recall that a cost-matrix $\Phi$ is said to be *reasonable* when $\phi_{ii} < \phi_{ji}$ for all $i, j$ with $j \neq i$ (see Section 2.1). As in Section 2.2 we distinguish between the estimation procedure $\hat{f}_n^{\text{est}} : \mathcal{Z}^n \to \mathcal{Y}^{\mathcal{X}}$ and the classifier $\hat{f}_n : \mathcal{X} \to \mathcal{Y}$, which implicitly depends on the data set $\mathcal{D}_n$. A key feature of our bound is that the constants are uniform over all manifolds $\mathcal{M}$ of a given reach $\tau$ and intrinsic dimension $\gamma$.

**Theorem 2** *Take $d \in \mathbb{N}$ and let $\rho$ be the Euclidean metric on $\mathbb{R}^d$. Suppose that $\Phi$ is a* reasonable *cost matrix and $\mathcal{M} \subseteq \mathbb{R}^d$ is a compact smooth submanifold with dimension $\gamma$, reach $\tau$ and Riemannian volume $V_{\mathcal{M}}$. There exists a positive constant $Z_\Phi$, determined by $\Phi$, and positive constants $C_0, R_0, V_-, V_+$ determined by $\gamma, \tau$, such that for all $c_0 \in (0, C_0)$, $r_0 \in (0, R_0)$, $\nu_{\min} \in (0, V_-)$, $\nu_{\max} \in (V_+, \infty)$, $\zeta_{\max} \in (0, Z_\Phi)$, $\alpha \in (0,1)$, $\beta \in (0, \gamma/\alpha)$, $C_\alpha, C_\beta > 0$, if we take $\Gamma = \langle (c_0, r_0, \nu_{\min}, \nu_{\max}), (\beta, C_\beta, \zeta_{\max}), (\alpha, C_\alpha) \rangle$ we have*

$$\inf \left\{ \sup \left\{ \mathbb{E}_{\mathbb{P}_n} \left[ R\left(\hat{f}_n\right)\right] - R^* : \mathbb{P} \in \mathcal{P}_\Phi\left(V_{\mathcal{M}}, \Gamma\right) \right\} : \hat{f}_n^{est} \in \left(\mathcal{Y}^{\mathcal{X}}\right)^{\mathcal{Z}^n} \right\} = \Theta\left( n^{-\frac{\alpha(1+\beta)}{2\alpha+\gamma}} \right),$$

*with upper and lower constants determined solely by $\Phi$, $\gamma$, $\tau$ and $\Gamma$.*

The proof of Theorem 2 consists of a a lower bound and an upper bound. The proof strategy for the lower bound is based on the proof of (Audibert et al., 2007, Theorem 3.5) where the result is proved in the special setting of binary classification with the zero-one loss $\Phi_{0,1}$ and a Lebesgue absolutely continuous marginal distribution $\mu$. In the proof of (Audibert et al., 2007, Theorem 3.5) the authors exploit the following fact: Suppose that you are given one of a pair of Bernoulli measures with probability $p = 1/2 + \Delta$ or $p = 1/2 - \Delta$ for some $\Delta \in (0, 1/2)$. If you know which of the two measures you have ($p = 1/2 + \Delta$ or $p = 1/2 - \Delta$) then you may make binary predictions in such a way as to get an error rate of $1/2 - \Delta$. However, without knowledge of which Bernoulli measure you have, your average expected error rate must be $1/2$, by symmetry. Hence, the average level of *regret* due to not knowing which of the two measures you have is $\Delta/2$. With this in mind, the structure of Euclidean space is then utilised to show that the set $[0, 1]^d$ may be broken up into $2^{qd}$ small well-separated cubes of size $2^{-qd-1}$. Thus, one can construct a set of measures $\mathbb{P}$ with a marginal $\mu$ supported on those cubes, and conditional equal to $p_\pm = 1/2 \pm \Delta$, with the choice of $\pm$ made independently for each small cube. The number of cubes means that a large number of training examples is required for an estimator to know the true value of the conditional on many of the small cubes. Hence, on average, an estimator based on $\mathcal{D}_n$ must have regret at least $\Delta/2$, on a large proportion of the cubes. This implies a lower bound on the generalisation error minus the Bayes error. In the proof of the lower bound for our Theorem 2 we have two important differences requiring modifications to the proof. Firstly, we are working in a multi-class cost sensitive scenario. Secondly, we are working on an embedded Riemannian manifold, rather than Euclidean space. To deal with the fact that our problem is multi-class and cost-sensitive we make use of Elkan's reasonableness assumption (Elkan (2001)) to show that there exists families of measures on the simplex $\Delta(\mathcal{Y})$ such that whatever class one predicts, the average expected cost one incurs, averaged over all the measures, is well separated from the expected cost one would have incurred if one knew which measure in $\Delta(\mathcal{Y})$ one had (see Lemmas A.1 and A.2). The key difficulty in the non-Euclidean setting is the construction of distributions which are both $(c_0, r_0)$ and satisfy the margin condition. Our construction consists a collection of well-separated closed geodesic balls (see Lemmas A.3 and A.5). To show that the support of the marginal is $(c_0, r_0)$-regular we exploit two key properties which follow from the assumption of bounded reach. Firstly, by results of Eftekhari and Wakin (2015) and Chazal (2013) the volume of small geodesic balls in manifolds of bounded reach is approximately $r^\gamma$ (see Lemma A.6). Secondly, we have a lower bound on the volume of the intersection of two sufficiently close geodesic balls (see Lemma A.7). We combine these properties to construct families of measures for which the average difference between expected risk and Bayes risk is bounded from below for all estimators. The full proof of the lower bound is given in Section A. The upper bound follows from Theorem 3 in Section 5 where we exhibit an efficient algorithm which attains the optimal rate.

## 5. The approximate nearest neighbour method on manifolds

In this section we shall see that the minimax rates identified in Section 4 are attained by the approximate nearest neighbour method. Given a data set $\mathcal{D}_n \sim \mathbb{P}^n$, an approximate nearest neighbour generating process $S = \{S_k\}_{k \in \mathbb{N}}$ and a query point $x \in \mathcal{X}$, the algorithm proceeds as follows:

1. Compute a set of approximate $k$-nearest neighbours $S_k(x, \mathcal{F}_n)$,
2. Estimate $\eta(x)$ with $\hat{\eta}^S_{n,k}(x) = \frac{1}{k} \sum_{i \in S_k(x, \mathcal{F}_n)} e(Y_i)$,
3. Predict $f^S_{n,k}(x) \in \text{argmin}_{y \in \mathcal{Y}} \left\{ e(y)^T \, \Phi \, \hat{\eta}^S_{n,k}(x) \right\}$.

The following result implies that the approximate nearest neighbour method is minimax optimal. To state the result we introduce the quantities of $\text{Asym}(\Phi)$ and $\Lambda(\Phi)$ which depend upon the cost-matrix $\Phi$. Given a cost matrix $\Phi$ we let $\text{Asym}(\Phi)$ denote the asymmetry of $\Phi$, given by $\text{Asym}(\Phi) := \max\{|\phi_{i_0 j} - \phi_{i_1 j}| : i_0 \neq j, i_1 \neq j\}$. Note that with $\Phi_{0-1}$ equal to the cost matrix corresponding to the zero-one loss we have $\text{Asym}(\Phi_{0-1}) = 0$. In addition, we define

$$\Lambda(\Phi) := (L - 2) \cdot \text{Asym}(\Phi) + 2\|\Phi\|_\infty.$$

The constant $\Lambda(\Phi)$ controls the degree of dependence of the cost differentials between classes, upon the marginal: The greater $\Lambda(\Phi)$ is, the greater the potential for errors in estimating $\eta$ to translate into incorrect assignments of relative cost.

**Theorem 3** *Take $d \in \mathbb{N}$ and let $\rho$ denote the Euclidean metric on $\mathbb{R}^d$. Let $\Phi$ be a cost matrix and $\mathcal{M} \subseteq \mathbb{R}^d$ a compact smooth submanifold with dimension $\gamma$ and reach $\tau$. Take positive constants $k_0, r_0, c_0, \nu_{\min}, \nu_{\max}, \zeta_{\max}, \alpha, \beta, C_\alpha, C_\beta$ and let $\Gamma = \langle (c_0, r_0, \nu_{\min}, \nu_{\max}), (\beta, C_\beta, \zeta_{\max}), (\alpha, C_\alpha) \rangle$. Suppose that $S$ generates $\theta$-approximate nearest neighbours for some $\theta \geq 1$. There exists a constant $C > 0$, depending upon $k_0$, $\gamma$, $\tau$, $\Gamma$ such that for all $\mathbb{P} \in \mathcal{P}_\Phi(V_\mathcal{M}, \Gamma)$ and $n \in \mathbb{N}$ the following holds:*

*(1) Given $\xi \in (0,1)$ and $k_n = k_0 \cdot n^{\frac{2\alpha}{2\alpha+\gamma}} \cdot (1 + \log(1/\xi))^{\gamma/(2\alpha+\gamma)}$ with probability at least $1 - \xi$ over $\mathcal{D}_n \sim \mathbb{P}^n$ we have*

$$\mathbb{P}\left[f_{n,k}^S(X) \notin \mathcal{Y}_\Phi^*(\eta(X))\right] \leq \xi + C \cdot \left(\theta^\alpha \cdot \Lambda(\Phi) \cdot \sqrt{\log(L)}\right)^\beta \cdot \left(\frac{1 + \log(1/\xi)}{n}\right)^{\beta\alpha/(2\alpha+\gamma)}.$$

*(2) Given $k_n = k_0 \cdot n^{\frac{2\alpha}{2\alpha+\gamma}}$ we have*

$$\mathbb{E}_n\left[R\left(f_{n,k_n}^S\right)\right] - R^* \leq C \cdot (\theta^\alpha \cdot \Lambda(\Phi))^{1+\beta} \cdot L \cdot n^{-\frac{\alpha(1+\beta)}{2\alpha+\gamma}}.$$

*Moreover, there exists an absolute constant $K > 0$ such that whenever $\theta > 1$, given any subgaussian random projection $\varphi : \mathbb{R}^d \to \mathbb{R}^h$ with*

$$h \geq K \cdot \|\varphi\|_{\psi_2}^4 \cdot \left(\frac{\theta^2+1}{\theta^2-1}\right)^2 \cdot \max\left\{\gamma \log_+(\gamma/(r_0 \cdot \tau)) - \log_+(c_0 \cdot \nu_{\min}) + \gamma, \log \delta^{-1}\right\},$$

*with probability at least $1 - \delta$, $S(\varphi)$ generates $\theta$-approximate nearest neighbours, so both (1) and (2) hold with $f_{n,k}^\varphi$ in place of $f_{n,k}^S$.*

We emphasize that the rates are uniform over all manifolds $\mathcal{M}$ of a given reach $\tau$ and intrinsic dimension $\gamma$ (for fixed $k_0, \Gamma$). Note that the first part of Theorem 3 includes the special case in which $\theta = 1$ and $S$ generates exact nearest neighbours. Approximate nearest neighbours and random projections are required purely for reducing computational complexity, and not for generalization performance. Note also that Theorem 3 holds for all non-negative cost matrices $\Phi$ (not necessarily reasonable).

A full proof of Theorem 3 is given in Section D. The first part of Theorem 3 follows straightforwardly from a more general result for metric spaces, which we present in Section 6. The second part of Theorem 3 follows from the first part combined with the following result on random projections. A full proof is given in Section D.

**Theorem 4** *There exists an absolute constant $K$ such that the following holds. Given a compact smooth submanifold $\mathcal{M} \subseteq \mathbb{R}^d$ with dimension $\gamma$ and reach $\tau$, suppose that $A \subset \mathcal{M}$ is $(c_0, r_0)$ regular with respect to the Riemannian volume $V_\mathcal{M}$. Suppose that $\varphi : \mathbb{R}^d \to \mathbb{R}^h$ is a subgaussian random projection. Take $\epsilon, \delta \in (0,1)$ and suppose that*

$$h \geq K \cdot \|\varphi\|_{\psi_2}^4 \cdot \epsilon^{-2} \cdot \max\left\{\gamma \log_+(\gamma/(r_0 \cdot \tau)) + \log_+(V_\mathcal{M}(A)/c_0) + \gamma, \log \delta^{-1}\right\}.$$

*Then with probability at least $1 - \delta$, for all pairs $x_0, x_1 \in A$ we have*

$$(1-\epsilon) \cdot \|x_0 - x_1\|_2^2 \leq \|\varphi(x_0) - \varphi(x_1)\|_2^2 \leq (1+\epsilon) \cdot \|x_0 - x_1\|_2^2.$$

Theorem 4 is a generalisation of (Dirksen, 2016, Theorem 7.9), where the result is given in the special case where $A = \mathcal{M}$. The proof is very similar, but is given in Section G for completeness. Theorem 4 is necessary for dealing with situations where we are able to bound the volume of the support of the marginal via the regularity condition (Defintion 2.1), but we have no bound on the volume of the ambient manifold $\mathcal{M}$.

## 6. The approximate nearest neighbour method on metric spaces

In this section we give a counterpart to Theorem 3 for arbitrary metric spaces. We first introduce the concept of *measure*-approximate nearest neighbours. Given $\omega \geq 1$, a family of mappings $S = \{S_k\}_{k\in\mathbb{N}}$ is said to generate $\omega$ measure-approximate nearest neighbours if, for each $x \in \mathcal{X}$, $n \in \mathbb{N}$ and $k \leq n$, $S_k(x, \mathcal{F}_n)$ is a subset of $\{1, \cdots, n\}$ with cardinality $k$ such that taking $r_0 = \max\{\rho(x, X_i) : i \in S_k^\circ(x, \mathcal{F}_n)\}$ and $r_1 = \max\{\rho(x, X_i) : i \in S_k(x, \mathcal{F}_n)\}$ implies $\mu(B_{r_1}(x)) \leq \omega \cdot \mu(B_{r_0}(x))$.

**Theorem 5** *Suppose that $\mathbb{P}$ satisfies the margin condition with constants $(\beta, C_\beta, \zeta_{\max})$ and that the conditional $\eta$ is measure-smooth, with constants $(\lambda, C_\lambda)$. Suppose that $S$ generates $\omega$ measure-approximate nearest neighbours with respect to the measure $\mu$ and take some $k_0 > 0$. There exists a constant $C > 0$, depending upon $k_0$, $\beta$, $\lambda$, $C_\beta$, $C_\lambda$ such that for all $n \in \mathbb{N}$ the following holds:*

*(1) Given $\xi \in (0, 1)$ and $k_n = k_0 \cdot n^{\frac{2\lambda}{2\lambda+1}} \cdot (1 + \log(1/\xi))^{1/(2\lambda+1)}$ with probability at least $1 - \xi$ over $\mathcal{D}_n \sim \mathbb{P}^n$ we have*

$$\mathbb{P}\left[f_{n,k}^S(X) \notin \mathcal{Y}_\Phi^*(\eta(X))\right] \leq \xi + C \cdot \left(\omega^\lambda \cdot \Lambda(\Phi) \cdot \sqrt{\log(L)}\right)^\beta \cdot \left(\frac{1 + \log(1/\xi)}{n}\right)^{\beta\lambda/(2\lambda+1)}.$$

*(2) Given $k_n = k_0 \cdot n^{\frac{2\lambda}{2\lambda+1}}$ we have*

$$\mathbb{E}_n\left[R\left(f_{n,k_n}^S\right)\right] - R^* \leq C \cdot \left(\omega^\lambda \cdot \Lambda(\Phi)\right)^{1+\beta} \cdot L \cdot n^{-\frac{\lambda(1+\beta)}{2\lambda+1}}.$$

Theorem 5 is an analogue of Theorem 4 in Chaudhuri and Dasgupta (2014), extended to the multi-class, cost-sensitive setting with measure approximate nearest neighbours. A sketch of the proof is as follows:

(1) By the concentration of measure phenomenon, given a data set $\mathcal{D}_n$ of size $n$ we expect the $k$ nearest neighbours of $x$ to lie in metric ball with probability roughly $k/n$, when $k$ is large. It follows that a set of $k$ $\omega$-measure-approximate nearest neighbours will with high probability lie in ball of probability roughly $\omega \cdot k/n$. By the measure smooth property (Definition 2.6) the conditional probability $\eta$ at those $k$ $\omega$-measure approximate nearest neighbours will be of the order $(\omega \cdot k/n)^\lambda$ from $\eta(x)$ or less, with high probability. By the margin condition, with high probability, the margin at $x$, $M_\Phi(x)$, is large. Moreover, if $x$ has large margin then the conditional of the $k$ $\omega$-measure approximate nearest neighbours would have to be far from $\eta(x)$ to lead to sub-optimal classifications, which is unlikely for small $(\omega \cdot k/n)^\lambda$. A more precise statement of this argument gives the conclusion that the predictions are optimal with high probability.

(2) The argument for (2) is more involved. We begin with the straightforward observation that the difference between expected risk of the approximate nearest neighbour classifier and the Bayes risk is equal to the average differential between the cost incurred by the approximate nearest neighbour classifier and the cost incurred by the Bayes optimal classifier. Let's denote this differential by $d(x, \mathcal{D}_n)$. The idea is to slice the difference between expected risk and Bayes risk up into regions based upon the value of $d(x, \mathcal{D}_n)$. For each $j$ we consider the event $d(x, \mathcal{D}_n) \in \left(2^{j-1} \cdot \epsilon, 2^j \cdot \epsilon\right]$. We note the following: a) By definition $d(x, \mathcal{D}_n) < 2^j \cdot \epsilon$ on the $j$-th slice, b) We have $0 < M_\Phi(\eta(x)) \leq d(x, \mathcal{D}_n) \leq 2^j \cdot \epsilon$, so we can upper bound the margin on the slice, c) The fact that $d(x, \mathcal{D}_n) > 2^{j-1} \cdot \epsilon$ implies that conditional of the approximate nearest neighbours is far from that of the test point. Both b) and c) are low probability events, so we can upper bound the probability of that $d(x, \mathcal{D}_n) \in \left(2^{j-1} \cdot \epsilon, 2^j \cdot \epsilon\right]$, and by a) we can upper bound the differential $d(x, \mathcal{D}_n)$, conditioned on the slice. Summing over the different slices gives an upper bound on the average value of $d(x, \mathcal{D}_n)$. By our initial observation, this translates into an upper bound on the difference between expected risk of the approximate nearest neighbour classifier and the Bayes risk. The full proof of Theorem 5 is given in Sections B and C.

## 7. Discussion

In this work we determined the minimax learning rates for a natural family of distributions supported on embedded manifolds, in a cost-sensitive setting. We proved that these rates are attained by an approximate nearest neighbour algorithm, where neighbours are computed in a low-dimensional randomly projected space. We also gave a bound upon on the number of dimensions required for the projection to attain optimal learning rates. Our work raises many questions for future investigation, both theoretical and empirical. Firstly, whilst we have demonstrated that Theorem 3 is optimal in the number of examples, up to a constant term, the bound depends linearly on the number of classes $L$, even in the cost-symmetric case. Whilst this is superior to the quadratic dependence of Crammer and Singer (2002), the distribution-free bounds of Kontorovich and Weiss (2014) give $O(\sqrt{\log(L)/n})$ dependence on the number of classes. It would be interesting to see if the approach of Kontorovich and Weiss (2014) may be adapted to give a better dependency upon $n$, in the presence of the margin condition. Secondly, as discussed in Section 2 Audibert and Tsybakov also considered higher order smoothness conditions and showed that in the presence of such conditions even faster learning rates are attainable in the Lebesgue absolutely continuous setting (see (Audibert et al., 2007, Section 2 & Section 3, Theorem 3.3)). In future work we intend to prove analogous results for manifolds, extending Theorem 2 to the setting of higher order smoothness conditions. From a more geometric perspective it would be interesting to see what bounds are possible if we relax the assumption that the data is concentrated on a manifold, and instead assume that the data lies near the manifold. The minimax optimality of the randomly projected nearest neighbour method in the cost-sensitive settings strongly suggests the method as a simple baseline for cost-sensitive problems. Hence, it would interesting to conduct an empirical investigation to determine how well the method compares on real-world data sets with other approaches to cost-sensitive classification Dmochowski et al. (2010); Nikolaou et al. (2016).

## Appendix A. Lower bound for cost-sensitive learning on manifolds

In order to prove Theorem 2 we first prove the following lower bound.

**Proposition A.1** *Suppose that $\Phi$ is a cost matrix satisfying Elkan's reasonableness assumption and $\mathcal{M} \subseteq \mathbb{R}^d$ is a compact smooth submanifold with dimension $\gamma$ and reach $\tau$. Let $\tilde{\tau} = \min\{\tau, 1\}$ and let $v_\gamma$ denote the volume of the $\gamma$-dimensional Euclidean unit ball. There exists a positive constant $Z_\Phi$, determined by $\Phi$, such that for all $c_0 \in \left(0, 2^{-14\gamma}\right)$, $r_0 \in (0, \tilde{\tau}/16)$, $\nu_{\min} \in \left(0, (2/\tilde{\tau})^\gamma \cdot v_\gamma^{-1}\right)$, $\nu_{\max} \in \left((2^{14}/\tilde{\tau})^\gamma \cdot v_\gamma^{-1}, \infty\right)$, $\zeta_{\max} \in (0, Z_\Phi)$, $\alpha \in (0,1)$, $\beta \in (0, \gamma/\alpha)$, $C_\alpha, C_\beta > 0$, if we take $\Gamma = \langle (c_0, r_0, \nu_{\min}, \nu_{\max}), (\beta, C_\beta, \zeta_{\max}), (\alpha, C_\alpha) \rangle$ then there exists a constant $C$ determined solely by $\Phi$, $\gamma$, $\tau$ and $\Gamma$ such that for all estimators $\hat{f} : \mathcal{Z}^n \to \mathcal{Y}^{\mathcal{X}}$ and $n \in \mathbb{N}$ we have*

$$\sup \left\{ \mathbb{E}_{\mathbb{P}_n} \left[ R\left(\hat{f}_n\right)\right] - R^* : \mathbb{P} \in \mathcal{P}_\Phi\left(V_{\mathcal{M}}, \Gamma\right) \right\} \geq C \cdot n^{-\frac{\alpha(1+\beta)}{2\alpha+\gamma}}.$$

To clarify the proof of Proposition A.1 requires several lemmas so for clarity we begin with an outline. First, some notation: Given a class label $y \in \mathcal{Y}$ and a probability vector $\boldsymbol{n} \in \Delta(\mathcal{Y})$ we shall define

$$D_\Phi(y, \boldsymbol{n}) = e(y)^T \Phi\, \boldsymbol{n} - \min_{l \in \mathcal{Y}} \left\{ e(l)^T \Phi\, \boldsymbol{n} \right\}.$$

Hence, $D_\Phi(y, \boldsymbol{n})$ is a form of *regret* which quantifies how far a prediction $y$ is from the optimal, according to a distribution $\boldsymbol{n}$ over $\mathcal{Y}$. The proof proceeds as follows:

**Lemma A.1:** We make use of Elkan's reasonableness assumption to construct a pair of measures in the simplex $\Delta(\mathcal{Y})$ such that 1) Both measures have large margin, and 2) For any predicted class label, the regret $D_\Phi(y, \boldsymbol{n})$ for at least one of the measures must be large.

**Lemma A.2:** This lemma is based on Assoud's lemma (Tsybakov, 2009, Chapter 2) and is closely related to the original construction within Audibert et al. (2007). We construct a family of measures $\mathbb{P}_{\boldsymbol{\sigma}}$ on $\mathcal{Z} = \mathcal{X} \times \mathcal{Y}$ such that we can lower bound the average differential between the expected risk and the Bayes risk. This lower bound is carried out by showing that with large probability $\mu$ on $x \in \mathcal{X}$, the corresponding conditionals $\eta_{\boldsymbol{\sigma}}(x)$ for measures $\mathbb{P}_{\boldsymbol{\sigma}}$ correspond to those constructed in Lemma A.1. Hence, by Lemma A.1 the average regret $D_\Phi\left(y, \eta_{\boldsymbol{\sigma}}(x)\right)$ for any prediction $y \in \mathcal{Y}$ is large. In addition it is shown that the different distributions $\mathbb{P}_{\boldsymbol{\sigma}}$ within the family are sufficiently similar that they cannot be effectively distinguished by any estimator $\hat{f}_n$ based on the data set $\mathcal{D}_n$.

**Lemma A.3:** We apply Lemma A.1 to give suitable conditions under which the family of measures constructed in Lemma A.3 satisfies the margin condition.

**Lemma A.4:** We give suitable conditions under which the family of measures constructed in Lemma A.3 satisfies the Hölder condition.

**Lemma A.7:** We use the geodesic structure of the manifold to show that when the centre of two metric balls are sufficiently close then the volume of their intersection is large.

**Lemma A.5:** We apply Lemma A.7 to construct $(c_0, r_0)$-regular sets $S(r)$ consisting of many small metric balls of radius $r$.

**Lemma A.9:** We give upper and lower bounds on the volume of the sets $S(r)$. This implies upper and lower bounds on the density of the normalised probability measure on $S(r)$.

**Lemma A.8:** We give a lower bound on the number of small balls in the supporting set $S(r)$.

Finally, Lemmas A.1-A.8 are combined. Using Lemma A.2, a family of measures is constructed with a lower bound on the difference between the risk of any estimator and the Bayes risk, averaged over all the measures. We then use Lemma A.3 to show that each of the measures in the family satisfies the margin condition and use Lemma A.4 to show that each measure has a sufficiently Holder conditional. Similarly, we use Lemmas A.5 and A.9 to show that the measure is $(c_0, r_0, \nu_{\min}, \nu_{\max})$ regular. Finally, Lemma A.8 is combined with other properties of the construction to show that the lower bound in Lemma A.2 implies the lower bound in Proposition A.1.

Given $p \in [0,1]$ we let $\boldsymbol{n}(p)$ denote the probability vector $\boldsymbol{n}(p) = (1-p)\cdot e(1) + p \cdot e(2)$.

**Lemma A.1** *Let $\Phi$ be a cost matrix with $L \geq 2$, satisfying Elkan's reasonableness assumption. There exists a constants $\kappa_\Phi \in (0,1)$, $t_\Phi \in (0, \min\{\kappa_\Phi, 1-\kappa_\Phi\})$ and $c_\Phi > 0$, depending solely upon $\Phi$, such that for all $\sigma \in \{-1,+1\}$, $\delta \in (0, t_\Phi)$ and all $y \in \mathcal{Y}$ we have*

$$\sum_{\sigma \in \{-1,+1\}} D_\Phi\left(y, \boldsymbol{n}\left(\kappa_\Phi + \sigma\cdot\delta\right)\right) \geq \min_{\sigma\in\{-1,+1\}} \left\{M_\Phi\left(\boldsymbol{n}\left(\kappa_\Phi + \sigma\cdot\delta\right)\right)\right\} \geq c_\Phi \cdot \delta.$$

**Proof** We begin by defining $\beta_\Phi = \min\left\{\phi_{y1} - \phi_{11} : y \in \mathcal{Y}\backslash\{1\}\right\}$,

$$\kappa_\Phi = \min\left\{\frac{(\phi_{y1} - \phi_{11})}{(\phi_{y1} - \phi_{11}) + (\phi_{12} - \phi_{y2})} : y \in \mathcal{Y}\backslash\{1\}, \;\; \phi_{y2} < \phi_{12}\right\},$$

and $c_\Phi = \beta_\Phi/(2\kappa_\Phi)$. By Elkan's reasonableness assumption we have $\phi_{y1} - \phi_{11} > 0$ for all $y \in \mathcal{Y}\backslash\{1\}$, so $\beta_\Phi > 0$ and $\phi_{22} < \phi_{12}$ so $\kappa_\Phi$ is well-defined and $\kappa_\Phi \in (0,1)$. We note that $\kappa_\Phi = \min\left\{\sup\left\{p \in (0,1) : e(y)^T\Phi\, \boldsymbol{n}(p) > e(1)^T\Phi\, \boldsymbol{n}(p)\right\} : y \in \mathcal{Y}\backslash\{1\}\right\}$. Thus, for all $p < \kappa_\Phi$ we have $\mathcal{Y}^*_\Phi\left(\boldsymbol{n}(p)\right) = \{1\}$ and for all $p > \kappa_\Phi$ we have $1 \notin \mathcal{Y}^*_\Phi\left(\boldsymbol{n}(p)\right)$. Hence, for all $\delta \in (0, \kappa)$ we have

$$\mathcal{Y}^*_\Phi\left(\boldsymbol{n}\left(\kappa_\Phi + \delta\right)\right) \cap \mathcal{Y}^*_\Phi\left(\boldsymbol{n}\left(\kappa_\Phi - \delta\right)\right) = \emptyset.$$

Thus,

$$\sum_{\sigma \in \{-1,+1\}} D_\Phi\left(y, \boldsymbol{n}\left(\kappa_\Phi + \sigma\cdot\delta\right)\right) \geq \min_{\sigma\in\{-1,+1\}} \left\{M_\Phi\left(\boldsymbol{n}\left(\kappa_\Phi + \sigma\cdot\delta\right)\right)\right\}.$$

Hence, it suffices to find some $t_\Phi \in (0, \min\{\kappa_\Phi, 1-\kappa_\Phi\})$ and $c_\Phi > 0$ such that for all $\delta \in (0, t_\Phi)$ we have

$$\min_{\sigma\in\{-1,+1\}} \left\{M_\Phi\left(\boldsymbol{n}\left(\kappa_\Phi + \sigma\cdot\delta\right)\right)\right\} \geq c_\Phi \cdot \delta,$$

We begin by showing that whenever $\delta < \kappa_\Phi$ and$y \in \mathcal{Y} \backslash \{1\}$ we have $D_\Phi(y, \boldsymbol{n}(\kappa_\Phi - \delta)) \geq (\beta_\Phi / \kappa_\Phi) \cdot \delta$.

Given $y \in \mathcal{Y} \backslash \{1\}$ we have $e(y)^T \Phi\, \boldsymbol{n}(0) - e(1)^T \Phi\, \boldsymbol{n}(0) \geq \beta_\Phi$ and $e(y)^T \Phi\, \boldsymbol{n}(\kappa_\Phi) - e(1)^T \Phi\, \boldsymbol{n}(\kappa_\Phi) \geq 0$. Hence, by mean value theorem together with the linearity of

$$p \mapsto \left(e(y)^T \Phi\, \boldsymbol{n}(p) - e(1)^T \Phi\, \boldsymbol{n}(p)\right) - \frac{\beta_\Phi}{\kappa_\Phi} \cdot (\kappa_\Phi - p),$$

we have

$$\begin{aligned} D_\Phi(y, \boldsymbol{n}(p)) &\geq \left(e(y)^T \Phi\, \boldsymbol{n}(p) - e(1)^T \Phi\, \boldsymbol{n}(p)\right) \\ &\geq \frac{\beta_\Phi}{\kappa_\Phi} \cdot (\kappa_\Phi - p). \end{aligned}$$

Hence, $D_\Phi(y, \boldsymbol{n}(\kappa_\Phi - \delta)) \geq (\beta_\Phi / \kappa_\Phi) \cdot \delta$ for all $\delta \in (0, \kappa_\Phi)$. Since this holds for all $y \in \mathcal{Y} \backslash \{1\}$ we have $M_\Phi(\boldsymbol{n}(\kappa_\Phi - \delta)) \geq (\beta_\Phi / \kappa_\Phi) \cdot \delta$ for all $\delta \in (0, \kappa_\Phi)$.

Now define sets $J_*, K_*, L_* \subset \mathcal{Y}$ by

$$\begin{aligned} J_* &= \{j \in \mathcal{Y} : \ (1 - \kappa_\Phi) \cdot (\phi_{j1} - \phi_{11}) + \kappa_\Phi \cdot (\phi_{j2} - \phi_{12}) = 0\}, \\ K_* &= \{j \in K_* : \ \phi_{j2} - \phi_{j1} \text{ is minimal}\}, \\ L_* &= \{j \in J_* \backslash K_* : \ \phi_{j2} - \phi_{j1} \text{ is minimal}\}. \end{aligned}$$

By the construction of $\kappa_\Phi$, we have $K_* \neq \emptyset$ and $1 \in J_* \backslash K_*$, so $L_* \neq \emptyset$. Since $\mathcal{Y}_\Phi^*(\boldsymbol{n}(p)) = \{1\}$ for all $p < \kappa_\Phi$, and for each $y \in \mathcal{Y}$, $e(y)^T \Phi \boldsymbol{n}(p)$ is linear in $p$ we have $J_* = \mathcal{Y}_\Phi^*(\boldsymbol{n}(\kappa_\Phi))$. Moreover, for each $y \in \mathcal{Y}$ we have

$$\frac{\partial \left(e(y)^T \Phi \boldsymbol{n}(p)\right)}{\partial p} = \phi_{j2} - \phi_{j1}.$$

Thus, there exists $t_\Phi \in (0, \min\{\kappa_\Phi, 1 - \kappa_\Phi\})$ such that for all $\delta \in (0, t_\Phi)$ we have $\mathcal{Y}_\Phi^*(\boldsymbol{n}(\kappa_\Phi + \delta)) = K_*$ and if we take $l \in L_*$ then for all $y \in \mathcal{Y} \backslash \mathcal{Y}_\Phi^*(\boldsymbol{n}(\kappa_\Phi + \delta))$ we have $e(l)^T \Phi \boldsymbol{n}(\kappa_\Phi + \delta) \leq e(y)^T \Phi \boldsymbol{n}(\kappa_\Phi + \delta)$. Choose $k_* \in K$ and $l_* \in L$. Then, given $\delta \in (0, t_\Phi)$, for all $y_0 \in \mathcal{Y}_\Phi^*(\boldsymbol{n}(\kappa_\Phi + \delta)) = K_*$ and $y_1 \in \mathcal{Y} \backslash \mathcal{Y}_\Phi^*(\boldsymbol{n}(\kappa_\Phi + \delta))$, we have

$$\begin{aligned} \left(e(y_1) - e(y_0)^T\right) \Phi\, \boldsymbol{n}(\kappa_\Phi + \delta) &\geq e(l^*)^T \Phi\, \boldsymbol{n}(\kappa_\Phi + \delta) - e(k^*)^T \Phi\, \boldsymbol{n}(\kappa_\Phi + \delta) \\ &= ((\phi_{l^*2} - \phi_{l^*1}) - (\phi_{k^*2} - \phi_{k^*1})) \cdot \delta. \end{aligned}$$

Thus, if we take $c_\Phi = \min\left\{\frac{\beta_\Phi}{\kappa_\Phi}, ((\phi_{l^*2} - \phi_{l^*1}) - (\phi_{k^*2} - \phi_{k^*1}))\right\} > 0$ then for all $\delta \in (0, t_\Phi)$,

$$\min_{\sigma \in \{-1, +1\}} \{M_\Phi(\boldsymbol{n}(\kappa_\Phi + \sigma \cdot \delta))\} \geq c_\Phi \cdot \delta.$$

∎

**Lemma A.2** *Let $\kappa_\Phi \in (0,1)$, $t_\Phi \in (0, \min\{\kappa_\Phi, 1 - \kappa_\Phi\})$ and $c_\Phi > 0$ be as in the statement of Lemma A.1. Fix a distribution $\mu$ on $\mathcal{X}$. Take $m \in \mathbb{N}$, together with positive constants $u \geq v > 0$. Suppose that we have two collections $\{A_j\}_{j=1}^m$, $\{B_j\}_{j=1}^m$ each consisting of $m$ disjoint subsets of*

*$\mathcal{X}$ such that for each $j \in \{1, \cdots, m\}$ we have $B_j \subset A_j$ and $v \leq \mu(B_j) \leq \mu(A_j) \leq u$. Suppose further that for each $j \in \{1, \cdots, m\}$ there exists a function $g_j : \mathcal{X} \to [0,1]$ with $g_j(x) = 0$ for all $x \notin A_j$ and $g_j(x) = 1$ for all $x \in B_j$. Take $\delta \in (0, t_\Phi)$ and for each $\boldsymbol{\sigma} \in \{-1, 0, +1\}^m$ we define $\eta_{\boldsymbol{\sigma}} : \mathcal{X} \to \mathbb{R}^{L \times 1}$ by*

$$\eta_{\boldsymbol{\sigma}}(x) = \boldsymbol{n}\left(\kappa_\Phi + \delta \cdot \sum_{j=1}^{m} \sigma_j \cdot g_j(x)\right),$$

*We let $\mathbb{P}_{\boldsymbol{\sigma}}$ denote the distribution on $\mathcal{Z} = \mathcal{X} \times \mathcal{Y}$ with marginal $\mu$ and conditional $\eta_{\boldsymbol{\sigma}}$. Let $\mathcal{P}$ be a set of distributions on $\mathcal{Z}$ with $\{\mathbb{P}_{\boldsymbol{\sigma}} : \boldsymbol{\sigma} \in \{-1, +1\}^m\} \subseteq \mathcal{P}$. Given any estimator $\hat{f} : \mathcal{Z}^N \to \mathcal{Y}^{\mathcal{X}}$ we have*

$$\sup_{\mathbb{P} \in \mathcal{P}} \left\{ \mathbb{E}_{\mathbb{P}^n}\left[ R_{\mathbb{P}}(\hat{f}_n) \right] - R_{\mathbb{P}}(f^*_{\mathbb{P}}) \right\} \geq (2 c_\Phi m v) \cdot \delta \cdot \left(1 - 2\delta \cdot \sqrt{\frac{nu}{t_\Phi}}\right).$$

**Proof** We have $\sup_{\mathbb{P} \in \mathcal{P}} \left\{ \mathbb{E}_{\mathbb{P}^n}\left[ R_{\mathbb{P}}(\hat{f}_n) \right] - R_{\mathbb{P}}(f^*_{\mathbb{P}}) \right\}$

$$\begin{aligned}
&= \sup_{\mathbb{P} \in \mathcal{P}} \left\{ \mathbb{E}_{\mathbb{P}^n}\left[ \mathbb{E}_{\mathbb{P}}\left[ \phi_{\hat{f}_n(X),Y} - \phi_{f^*_{\mathbb{P}}(X),Y} \right]\right]\right\} \\
&\geq \frac{1}{2^m} \sum_{\boldsymbol{\sigma} \in \{-1,+1\}^m} \mathbb{E}_{\mathbb{P}^n_{\boldsymbol{\sigma}}}\left[ \mathbb{E}_{\mathbb{P}_{\boldsymbol{\sigma}}}\left[ \phi_{\hat{f}_n(X),Y} - \phi_{f^*_{\mathbb{P}_{\boldsymbol{\sigma}}}(X),Y} \right]\right] \\
&= \frac{1}{2^m} \sum_{\boldsymbol{\sigma} \in \{-1,+1\}^m} \mathbb{E}_{\mathbb{P}^n_{\boldsymbol{\sigma}}}\left[ \int \left( e\left(\hat{f}_n(x)\right) - e\left(f^*_{\mathbb{P}_{\boldsymbol{\sigma}}}(x)\right)\right)^T \Phi \;\; \eta_{\boldsymbol{\sigma}}(x) d\mu(x) \right] \\
&= \frac{1}{2^m} \sum_{\boldsymbol{\sigma} \in \{-1,+1\}^m} \mathbb{E}_{\mathbb{P}^n_{\boldsymbol{\sigma}}}\left[ \int D_\Phi\left(\hat{f}_n(x), \eta_{\boldsymbol{\sigma}}(x)\right) d\mu(x) \right]
\end{aligned}$$

Note that $D_\Phi(y, \boldsymbol{n}) \geq 0$ for all $y \in \mathcal{Y}$ and probability vectors $\boldsymbol{n}$. Moreover, $\eta_{\boldsymbol{\sigma}}(x) = \boldsymbol{n}(\kappa_\Phi + \sigma_j \cdot \delta)$ for all $x \in B_j$. Hence, we have
$\sup_{\mathbb{P} \in \mathcal{P}} \left\{ \mathbb{E}_{\mathbb{P}^n}\left[ R_{\mathbb{P}}(\hat{f}_n) \right] - R_{\mathbb{P}}(f^*_{\mathbb{P}}) \right\}$

$$\begin{aligned}
&\geq \frac{1}{2^m} \sum_{\boldsymbol{\sigma} \in \{-1,+1\}^m} \mathbb{E}_{\mathbb{P}^n_{\boldsymbol{\sigma}}}\left[ \sum_{j=1}^{m} \int_{B_j} D_\Phi\left(\hat{f}_n(x), \eta_{\boldsymbol{\sigma}}(x)\right) d\mu(x) \right] \\
&\geq \sum_{j=1}^{m} \left( \frac{1}{2^m} \sum_{\boldsymbol{\sigma} \in \{-1,+1\}^m} \mathbb{E}_{\mathbb{P}^n_{\boldsymbol{\sigma}}}\left[ \int_{B_j} D_\Phi\left(\hat{f}_n(x), \kappa_\Phi + \sigma_j \cdot \delta\right) d\mu(x) \right]\right).
\end{aligned}$$

Given $j \in \{1, \cdots, m\}$, $\boldsymbol{\sigma} = (\sigma_i)_{i=1}^{m-1} \in \{-1, +1\}^{m-1}$ and $r \in \{-1, 0, +1\}$ define $(\boldsymbol{\sigma} \|_j r)$ by

$$(\boldsymbol{\sigma} \|_j r)_i = \begin{cases} \sigma_i & \text{if } i \in \{1, \cdots, j-1\} \\ r & \text{if } i = j \\ \sigma_{i-1} & \text{if } i \in \{j+1, \cdots, m\}. \end{cases}$$

From the above we have

$$\sup_{\mathbb{P}\in\mathcal{P}}\left\{\mathbb{E}_{\mathbb{P}^n}\left[R_{\mathbb{P}}(\hat{f}_n)\right]-R_{\mathbb{P}}(f^*_{\mathbb{P}})\right\}$$

$$\geq \sum_{j=1}^{m}\left(\frac{1}{2^m}\sum_{\boldsymbol{\sigma}\in\{-1,+1\}^{m-1}}\left(\sum_{r\in\{-1,+1\}}\mathbb{E}_{\mathbb{P}^n_{\left(\boldsymbol{\sigma}\|_j r\right)}}\left[\int_{B_j} D_\Phi\left(\hat{f}_n(x),\kappa_\Phi+r\cdot\delta\right)d\mu(x)\right]\right)\right)$$
$$\geq \sum_{j=1}^{m}\int_{B_j}\left(\left(\frac{1}{2}\right)^{m-1}\sum_{\boldsymbol{\sigma}\in\{-1,+1\}^{m-1}}\left(\frac{1}{2}\sum_{r\in\{-1,+1\}}\mathbb{E}_{\mathbb{P}^n_{\left(\boldsymbol{\sigma}\|_j r\right)}}\left[D_\Phi\left(\hat{f}_n(x),\kappa_\Phi+r\cdot\delta\right)\right]\right)\right)d\mu(x).$$

Hence, it suffices to fix $j\in\{1,\cdots,m\}$, $x\in B_j$, $\boldsymbol{\sigma}\in\{-1,+1\}^{m-1}$ and show that

$$\frac{1}{2}\sum_{r\in\{-1,+1\}}\mathbb{E}_{\mathbb{P}^n_{\left(\boldsymbol{\sigma}\|_j r\right)}}\left[D_\Phi\left(\hat{f}_n(x),\kappa_\Phi+r\cdot\delta\right)\right]\geq(2c_\Phi)\cdot\delta\cdot\left(1-2\delta\cdot\sqrt{\frac{nu}{t_\Phi}}\right).$$

For each $r\in\{-1,0,+1\}$ we let $\pi_{\boldsymbol{\sigma},j,r}$ denote the Radon-Nikodym derivative of $\mathbb{P}_{(\boldsymbol{\sigma}\|_j r)}$ with respect to $\mathbb{P}_{(\boldsymbol{\sigma}\|_j 0)}$. Similarly, we let $\pi^n_{\boldsymbol{\sigma},j,r}$ denote the Radon-Nikodym derivative of $\mathbb{P}^n_{(\boldsymbol{\sigma}\|_j r)}$ with respect to $\mathbb{P}^n_{(\boldsymbol{\sigma}\|_j 0)}$. From the definition of $\mathbb{P}_{(\boldsymbol{\sigma}\|_j r)}$ we have

$$\pi_{\boldsymbol{\sigma},j,r}\left((x,y)\right)=\begin{cases}1 \text{ if } x\notin A_j \text{ or } y\notin\{1,2\},\\ 1-r\cdot\delta\cdot g_j(x)/(1-\kappa_\Phi) \text{ if } x\in A_j \text{ and } y=1,\\ 1+r\cdot\delta\cdot g_j(x)/\kappa_\Phi \text{ if } x\in A_j \text{ and } y=2.\end{cases}\tag{1}$$

We have

$$\frac{1}{2}\sum_{r\in\{-1,+1\}}\mathbb{E}_{\mathbb{P}^n_{\left(\boldsymbol{\sigma}\|_j r\right)}}\left[D_\Phi\left(\hat{f}_n(x),\kappa_\Phi+r\cdot\delta\right)\right]$$

$$=\frac{1}{2}\sum_{r\in\{-1,+1\}}\mathbb{E}_{\mathbb{P}^n_{\left(\boldsymbol{\sigma}\|_j 0\right)}}\left[D_\Phi\left(\hat{f}_n(x),\kappa_\Phi+r\cdot\delta\right)\pi^n_{\boldsymbol{\sigma},j,r}\right]$$
$$\geq\mathbb{E}_{\mathbb{P}^n_{\left(\boldsymbol{\sigma}\|_j 0\right)}}\left[\left(\frac{1}{2}\sum_{r\in\{-1,+1\}}D_\Phi\left(\hat{f}_n(x),\kappa_\Phi+r\cdot\delta\right)\right)\cdot\min_{r\in\{-1,+1\}}\left\{\pi^n_{\boldsymbol{\sigma},j,r}\right\}\right]$$
$$\geq 2c_\Phi\cdot\delta\cdot\mathbb{E}_{\mathbb{P}^n_{\left(\boldsymbol{\sigma}\|_j 0\right)}}\left[\min_{r\in\{-1,+1\}}\left\{\pi^n_{\boldsymbol{\sigma},j,r}\right\}\right].$$

The final inequality follows from Lemma A.1. Hence, to complete the proof of the lemma it remains to show that

$$\mathbb{E}_{\mathbb{P}^n_{\left(\boldsymbol{\sigma}\|_j 0\right)}}\left[\min_{r\in\{-1,+1\}}\left\{\pi^n_{\boldsymbol{\sigma},j,r}\right\}\right]\geq 1-2\delta\cdot\sqrt{\frac{nu}{t_\Phi}}.$$

Equivalently, we must show that

$$\mathbb{E}_{\mathbb{P}^n_{(\boldsymbol{\sigma}\|_j 0)}}\left[\left|\pi^n_{\boldsymbol{\sigma},j,+1} - \pi^n_{\boldsymbol{\sigma},j,-1}\right|\right] \leq 2\delta \cdot \sqrt{\frac{nu}{t_\Phi}}, \tag{2}$$

where we have used the fact that

$$\mathbb{E}_{\mathbb{P}^n_{(\boldsymbol{\sigma}\|_j 0)}}\left[\pi^n_{\boldsymbol{\sigma},j,+1}\right] = \mathbb{E}_{\mathbb{P}^n_{(\boldsymbol{\sigma}\|_j 0)}}\left[\pi^n_{\boldsymbol{\sigma},j,-1}\right] = 1.$$

For each $\omega = (\omega_i)_{i=1}^n \in \{0,1\}^n$ we let

$$S_\omega := \left\{\mathcal{D}_n = (Z_i)_{i=1}^n : X_i \in A_j \text{ if and only if } \omega_i = 1\right\}.$$

Thus, we have $\mathbb{E}_{\mathbb{P}^n_{(\boldsymbol{\sigma}\|_j 0)}}\left[\left|\pi^n_{\boldsymbol{\sigma},j,+1} - \pi^n_{\boldsymbol{\sigma},j,-1}\right|\right]$

$$= \sum_{\omega\in\{0,1\}^n} \mathbb{P}^n_{(\boldsymbol{\sigma}\|_j 0)}\left[\mathcal{D}_n \in S_\omega\right] \cdot \mathbb{E}_{\mathbb{P}^n_{(\boldsymbol{\sigma}\|_j 0)}}\left[\left|\pi^n_{\boldsymbol{\sigma},j,+1} - \pi^n_{\boldsymbol{\sigma},j,-1}\right| |S_\omega\right]$$

We now bound $\mathbb{E}_{\mathbb{P}^n_{(\boldsymbol{\sigma}\|_j 0)}}\left[\left|\pi^n_{\boldsymbol{\sigma},j,+1} - \pi^n_{\boldsymbol{\sigma},j,-1}\right| |S_\omega\right]$ for each $\omega \in \{0,1\}^n$. We first deal with the simple case where $\sum_{i=1}\omega_i = 1$. Recall that $t_\Phi < (\min\{\kappa_\Phi, 1-\kappa_\Phi\})^{-1}$, so applying (1) gives

$$\begin{aligned}\mathbb{E}_{\mathbb{P}^n_{(\boldsymbol{\sigma}\|_j 0)}}\left[\left|\pi^n_{\boldsymbol{\sigma},j,+1} - \pi^n_{\boldsymbol{\sigma},j,-1}\right| |S_\omega\right] &= \mathbb{E}_{\mathbb{P}_{(\boldsymbol{\sigma}\|_j 0)}}\left[\left|\pi_{\boldsymbol{\sigma},j,+1} - \pi_{\boldsymbol{\sigma},j,-1}\right| |X \in A_j\right] \\ &\leq \frac{2}{t_\Phi}\cdot\delta \leq 2\delta\cdot\sqrt{\frac{\sum_{i=1}^n \omega_i}{t_\Phi}}.\end{aligned}$$

We now bound $\mathbb{E}_{\mathbb{P}^n_{(\boldsymbol{\sigma}\|_j 0)}}\left[\left|\pi^n_{\boldsymbol{\sigma},j,+1} - \pi^n_{\boldsymbol{\sigma},j,-1}\right| |S_\omega\right]$ for $\omega \in \{0,1\}^n$ with $\sum_{i=1}\omega_i \geq 2$.

Note that for each $r \in \{-1,0,+1\}$ we have

$$\mathbb{P}^n_{(\boldsymbol{\sigma}\|_j r)}\left[\mathcal{D}_n \in S_\omega\right] = \prod_{i=1}^n \mu\left(A_j\right)^{\omega_i} \cdot \left(1-\mu\left(A_j\right)\right)^{1-\omega_i}.$$

Hence,

$$\mathbb{E}_{\mathbb{P}^n_{(\boldsymbol{\sigma}\|_j 0)}}\left[\pi^n_{\boldsymbol{\sigma},j,+1}|S_\omega\right] = \mathbb{E}_{\mathbb{P}^n_{(\boldsymbol{\sigma}\|_j 0)}}\left[\pi^n_{\boldsymbol{\sigma},j,-1}|S_\omega\right] = 1.$$

Thus, by the Cauchy Schwartz inequality we have $\mathbb{E}_{\mathbb{P}^n_{(\boldsymbol{\sigma}\|_j 0)}}\left[\left|\pi^n_{\boldsymbol{\sigma},j,+1} - \pi^n_{\boldsymbol{\sigma},j,-1}\right| |S_\omega\right]$

$$\begin{aligned}&= \mathbb{E}_{\mathbb{P}^n_{(\boldsymbol{\sigma}\|_j 0)}}\left[\left(\sqrt{\pi^n_{\boldsymbol{\sigma},j,+1}} - \sqrt{\pi^n_{\boldsymbol{\sigma},j,-1}}\right)\left(\sqrt{\pi^n_{\boldsymbol{\sigma},j,+1}} + \sqrt{\pi^n_{\boldsymbol{\sigma},j,-1}}\right)|S_\omega\right] \\ &\leq \sqrt{\mathbb{E}_{\mathbb{P}^n_{(\boldsymbol{\sigma}\|_j 0)}}\left[\left(\sqrt{\pi^n_{\boldsymbol{\sigma},j,+1}} - \sqrt{\pi^n_{\boldsymbol{\sigma},j,-1}}\right)^2|S_\omega\right]} \cdot \sqrt{\mathbb{E}_{\mathbb{P}^n_{(\boldsymbol{\sigma}\|_j 0)}}\left[\left(\sqrt{\pi^n_{\boldsymbol{\sigma},j,+1}} + \sqrt{\pi^n_{\boldsymbol{\sigma},j,-1}}\right)^2|S_\omega\right]} \\ &= \sqrt{2\left(1 - \mathbb{E}_{\mathbb{P}^n_{(\boldsymbol{\sigma}\|_j 0)}}\left[\sqrt{\pi^n_{\boldsymbol{\sigma},j,+1}\cdot\pi^n_{\boldsymbol{\sigma},j,-1}}|S_\omega\right]\right)} \cdot \sqrt{2\left(1 + \mathbb{E}_{\mathbb{P}^n_{(\boldsymbol{\sigma}\|_j 0)}}\left[\sqrt{\pi^n_{\boldsymbol{\sigma},j,+1}}\cdot\sqrt{\pi^n_{\boldsymbol{\sigma},j,-1}}|S_\omega\right]\right)} \\ &= 2\sqrt{2}\cdot\sqrt{1 - \mathbb{E}_{\mathbb{P}^n_{(\boldsymbol{\sigma}\|_j 0)}}\left[\sqrt{\pi^n_{\boldsymbol{\sigma},j,+1}\cdot\pi^n_{\boldsymbol{\sigma},j,-1}}|S_\omega\right]}\end{aligned}$$

To bound this term we first note that for each $r \in \{-1, 0, +1\}$, $\mathbb{P}^n_{(\boldsymbol{\sigma}\|_j r)}$ is a product measure. Hence,

$$\mathbb{E}_{\mathbb{P}^n_{(\boldsymbol{\sigma}\|_j 0)}}\left[\sqrt{\pi^n_{\boldsymbol{\sigma},j,+1} \cdot \pi^n_{\boldsymbol{\sigma},j,-1}}|S_\omega\right] = \left(\mathbb{E}_{\mathbb{P}_{(\boldsymbol{\sigma}\|_j 0)}}\left[\sqrt{\pi_{\boldsymbol{\sigma},j,+1} \cdot \pi_{\boldsymbol{\sigma},j,-1}}|X \in A_j\right]\right)^{\sum_{i=1}^n \omega_i}$$
$$\cdot \left(\mathbb{E}_{\mathbb{P}_{(\boldsymbol{\sigma}\|_j 0)}}\left[\sqrt{\pi_{\boldsymbol{\sigma},j,+1} \cdot \pi_{\boldsymbol{\sigma},j,-1}}|X \notin A_j\right]\right)^{n-\sum_{i=1}^n \omega_i}$$

By construction, for all $x \notin A_j$ we have $\eta_{(\boldsymbol{\sigma}\|_j+1)}(x) = \eta_{(\boldsymbol{\sigma}\|_j+1)}(x) = \eta_{(\boldsymbol{\sigma}\|_j+1)}(x)$, so $\pi_{\boldsymbol{\sigma},j,+1}(x) = \pi_{\boldsymbol{\sigma},j,-1}(x) = \pi_{\boldsymbol{\sigma},j,0}(x)$. Hence,

$$\mathbb{E}_{\mathbb{P}_{(\boldsymbol{\sigma}\|_j 0)}}\left[\sqrt{\pi_{\boldsymbol{\sigma},j,+1} \cdot \pi_{\boldsymbol{\sigma},j,-1}}|X \notin A_j\right] = \mathbb{E}_{\mathbb{P}_{(\boldsymbol{\sigma}\|_j 0)}}\left[\pi_{\boldsymbol{\sigma},j,0}|X \notin A_j\right] = 1.$$

Consequently, we have

$$\mathbb{E}_{\mathbb{P}^n_{(\boldsymbol{\sigma}\|_j 0)}}\left[\sqrt{\pi^n_{\boldsymbol{\sigma},j,+1} \cdot \pi^n_{\boldsymbol{\sigma},j,-1}}|S_\omega\right] = \left(\mathbb{E}_{\mathbb{P}_{(\boldsymbol{\sigma}\|_j 0)}}\left[\sqrt{\pi_{\boldsymbol{\sigma},j,+1} \cdot \pi_{\boldsymbol{\sigma},j,-1}}|X \in A_j\right]\right)^{\sum_{i=1}^n \omega_i}$$

Moreover, by (1) we see that for all $(x, y) \in \mathcal{X} \times \mathcal{Y}$,

$$\pi_{\boldsymbol{\sigma},j,+1} \cdot \pi_{\boldsymbol{\sigma},j,-1} \geq 1 - \frac{r^2 \cdot \delta^2 \cdot g_j(x)}{t_\Phi} \geq 1 - \frac{\delta^2}{t_\Phi}.$$

Combining these inequalities we have,

$$\mathbb{E}_{\mathbb{P}^n_{(\boldsymbol{\sigma}\|_j 0)}}\left[\left|\pi^n_{\boldsymbol{\sigma},j,+1} - \pi^n_{\boldsymbol{\sigma},j,-1}\right| |S_\omega\right] \leq 2\sqrt{2} \cdot \sqrt{1 - \left(1 - \frac{\delta^2}{t_\Phi}\right)^{\frac{1}{2}\cdot\sum_{i=1}^n \omega_i}}.$$

Note that for any $l \geq 2$ and $x \geq 0$ we have $2\left(1 - \left(1 - x^2\right)^{\frac{l}{2}}\right) \leq lx^2$. Hence, we have

$$\mathbb{E}_{\mathbb{P}^n_{(\boldsymbol{\sigma}\|_j 0)}}\left[\left|\pi^n_{\boldsymbol{\sigma},j,+1} - \pi^n_{\boldsymbol{\sigma},j,-1}\right| |S_\omega\right] \leq 2\delta \cdot \sqrt{\frac{\sum_{i=1}^n \omega_i}{t_\Phi}}.$$

It follows that $\mathbb{E}_{\mathbb{P}^n_{(\boldsymbol{\sigma}\|_j 0)}}\left[\left|\pi^n_{\boldsymbol{\sigma},j,+1} - \pi^n_{\boldsymbol{\sigma},j,-1}\right|\right]$

$$\begin{aligned}
&= \sum_{\omega \in \{0,1\}^n} \mathbb{P}^n_{(\boldsymbol{\sigma}\|_j 0)}\left[\mathcal{D}_n \in S_\omega\right] \cdot \mathbb{E}_{\mathbb{P}^n_{(\boldsymbol{\sigma}\|_j 0)}}\left[\left|\pi^n_{\boldsymbol{\sigma},j,+1} - \pi^n_{\boldsymbol{\sigma},j,-1}\right| |S_\omega\right] \\
&\leq \frac{2\delta}{\sqrt{t_\Phi}} \cdot \sum_{\omega \in \{0,1\}^n} \prod_{i=1}^n \mu\left(A_j\right)^{\omega_i} \cdot \left(1 - \mu\left(A_j\right)\right)^{1-\omega_i} \cdot \sqrt{\sum_{i=1}^n \omega_i} \\
&\leq \frac{2\delta}{\sqrt{t_\Phi}} \cdot \sqrt{\sum_{\omega \in \{0,1\}^n} \prod_{i=1}^n \mu\left(A_j\right)^{\omega_i} \cdot \left(1 - \mu\left(A_j\right)\right)^{1-\omega_i} \cdot \sum_{i=1}^n \omega_i} \\
&= \frac{2\delta}{\sqrt{t_\Phi}} \cdot \sqrt{n \cdot \mu\left(A_j\right)} \leq 2\delta \cdot \sqrt{\frac{nu}{t_\Phi}}.
\end{aligned}$$

This completes the proof of the lemma. ■

**Lemma A.3** *Suppose that $\Phi$ satisfies Elkan's reasonableness assumption and let $c_\Phi, \kappa_\Phi, t_\Phi$ be as in the statement of Lemma A.1. Take $m \in \mathbb{N}$ and $u > 0$ and suppose that there are disjoint sets $\{A_j\}_{j=1}^m$ such that for each $j \in \{1, \cdots, m\}$ there exists $B_j \subset A_j$ with $\mu(B_j) \leq u$ and $\mu(A_j \backslash B_j) = 0$ along with a function $g_j : \mathcal{X} \to [0,1]$ with $g_j(x) = 0$ for all $x \notin A_j$ and $g_j(x) = 1$ for all $x \in B_j$. We also take $\boldsymbol{\sigma} \in \{-1, 0, +1\}^m$ along with $\eta_{\boldsymbol{\sigma}}$ and $\mathbb{P}_{\boldsymbol{\sigma}}$ as in the statement of Lemma A.2. Suppose we have constants $C_\beta > 0$, $\beta > 0$, $\zeta_{\max} \in (0, M_\Phi(\boldsymbol{n}(\kappa_\Phi)))$, and $\delta \in (0, t_\Phi)$ satisfying $m \cdot u \leq C_\beta \cdot (c_\Phi \cdot \delta)^\beta$. Then the measure $\mathbb{P}_{\boldsymbol{\sigma}}$ satisfies the margin condition with constants $(\beta, C_\beta, \zeta_{\max})$.*

**Proof** First note that by the construction of $\eta_{\boldsymbol{\sigma}}$ (see Lemma A.2), combined with the fact that if $x \in B_{j_x}$ we have $g_{j_x}(x) = 1$ and $g_j(x) = 0$ for $j \neq j_x$, so

$$\eta_{\boldsymbol{\sigma}}(x) = \boldsymbol{n}\left(\kappa_\Phi + \delta \cdot \sum_{j=1}^m \sigma_j \cdot g_j(x)\right) = \boldsymbol{n}\left(\kappa_\Phi + \delta \cdot \sigma_{j_x}\right).$$

Hence, by Lemma A.1 we have $M_\Phi(\eta_{\boldsymbol{\sigma}}(x)) \geq c_\Phi \cdot \delta$. On the other hand, if $x \in \mathcal{X} \backslash \bigcup_{j=1}^m A_j$ then $g_j(x) = 0$ for all $j$, so $\eta_{\boldsymbol{\sigma}}(x) = \boldsymbol{n}(\kappa_\Phi)$, so $M(\eta_{\boldsymbol{\sigma}}(x)) > \zeta_{\max}$.

We now fix $\zeta < \zeta_{\max} < M_\Phi(\boldsymbol{n}(\kappa_\Phi))$. We must show that,

$$\mu(\{x \in \mathcal{X} : M_\Phi(\eta(x)) \leq \zeta\}) \leq C_\beta \cdot \zeta^\beta.$$

Now if $\zeta < c_\Phi \cdot \delta$ then

$$\mu(\{x \in \mathcal{X} : M_\Phi(\eta(x)) \leq \zeta\}) \leq \mu\left(\bigcup_{j=1}^m A_j \backslash B_j\right) = 0.$$

On the other hand, if $\zeta \in (c_\Phi \cdot \delta, \zeta_{\max})$, then

$$\begin{aligned}\mu(\{x \in \mathcal{X} : M_\Phi(\eta(x)) \leq \zeta\}) &\leq \mu\left(\bigcup_{j=1}^m A_j\right) \\ &\leq m \cdot u \leq C_\beta \cdot (c_\Phi \delta)^\beta \leq C_\beta \cdot \zeta^\beta.\end{aligned}$$

■

**Lemma A.4** *Suppose that $\mathcal{W} = \{w_j\}_{j=1}^m$ is a $r$-separated set for some $r > 0$. There exist functions $\{g_j\}_{j=1}^m$ such that for each $j \in \{1, \cdots, m\}$, $g_j(x) = 0$ for all $x \notin B_{r/3}(w_j)$, $g_j(x) = 1$ for all $x \in B_{r/6}(w_j)$ and for any $C_\alpha > 0$, $\alpha \in (0, 1]$, $\delta \in (0, (C_\alpha/12) \cdot r^\alpha]$ and $\boldsymbol{\sigma} \in \{-1, 0, +1\}^m$, the function $\eta_{\boldsymbol{\sigma}} : \mathcal{X} \to \mathbb{R}^{L \times 1}$ by*

$$\eta_{\boldsymbol{\sigma}}(x) = \boldsymbol{n}\left(\kappa_\Phi + \delta \cdot \sum_{j=1}^m \sigma_j \cdot g_j(x)\right),$$

*is Hölder continuous with constants $(\alpha, C_\alpha)$.*

**Proof** Firstly, we define a function $u : [0,1] \to [0,1]$ by

$$u(t) = \begin{cases} 1 & \text{for } t \in [0, 1/3] \\ 2 - 3t & \text{for } t \in [1/3, 2/3] \\ 0 & \text{for } t \geq 2/3. \end{cases}$$

For each $j \in \{1, \cdots, m\}$ we let $g_j(x) = u\left((2/r) \cdot \rho(x, w_j)\right)$. Clearly if $x \notin B_{r/3}(w_j)$ then $(2/r) \cdot \rho(x, w_j) \geq 2/3$ so $g_j(x) = 0$. On the other hand, if $x \in B_{r/6}(w_j)$ then $(2/r) \cdot \rho(x, w_j) < 1/3$, so $g_j(x) = 1$. We now fix $\delta \in (0, (C_\alpha/12) \cdot r^\alpha]$ and $\boldsymbol{\sigma} \in \{-1, 0, +1\}^m$ and show that $\eta_{\boldsymbol{\sigma}}$ is Hölder continuous with constants $(\alpha, C_\alpha)$. By the definition of $\boldsymbol{n}$ it suffices to show that

$$\varphi_{\boldsymbol{\sigma}}(x) = \kappa_\Phi + \delta \cdot \sum_{j=1}^{m} \sigma_j \cdot g_j(x)$$

is Hölder continuous with constants $(\alpha, C_\alpha)$. Since $g_j(x) = 0$ for all $x \notin B_{r/3}(w_j)$ and $\{w_j\}_{j=1}^m$ is an $r$-separated set, we have $\varphi_{\boldsymbol{\sigma}}(x) = \delta \cdot \sigma_{j_x} \cdot u\left((2/r) \cdot \rho(x, w_{j_x})\right)$ whenever $x \in B_{r/3}(w_{j_x})$ for some $j_x \in \{1, \cdots, m\}$, and $\varphi(x) = 0$ if $x \in \mathcal{X} \backslash \bigcup_{j=1}^m B_{r/3}(w_j)$. Now take $x_0, x_1 \in \mathcal{X}$. If $\varphi_{\boldsymbol{\sigma}}(x_0) = \varphi_{\boldsymbol{\sigma}}(x_1) = 0$ then $\|\varphi_{\boldsymbol{\sigma}}(x_0) - \varphi_{\boldsymbol{\sigma}}(x_1)\| \leq C_\alpha \cdot \rho(x_0, x_1)^\alpha$ holds trivially. Now suppose that $\varphi_{\boldsymbol{\sigma}}(x_0) \neq 0$ or $\varphi_{\boldsymbol{\sigma}}(x_1) \neq 0$. Without loss of generality we assume that $\varphi_{\boldsymbol{\sigma}}(x_0) \neq 0$. Hence, for some $j_0 \in \{1, \cdots, m\}$ we have $x_0 \in B_{r/3}(w_{j_0})$. Now either $x_1 \in B_{r/2}(w_{j_0})$ or $x_1 \notin B_{r/2}(w_{j_0})$. If $x_1 \in B_{r/2}(w_{j_0})$ then we have $\varphi_{\boldsymbol{\sigma}}(x_0) = \delta \cdot \sigma_{j_0} \cdot u\left((2/r) \cdot \rho(x_0, w_{j_0})\right)$ and $\varphi_{\boldsymbol{\sigma}}(x_1) = \delta \cdot \sigma_{j_0} \cdot u\left((2/r) \cdot \rho(x_1, w_{j_0})\right)$. Moreover, $|\rho(x_0, w_{j_0}) - \rho(x_1, w_{j_0})| \leq \rho(x_0, x_1)$, so $|u\left((2/r) \cdot \rho(x_0, w_{j_0})\right) - u\left((2/r) \cdot \rho(x_1, w_{j_0})\right)| \leq (6/r) \cdot \rho(x_0, x_1)$. Hence,

$$\begin{aligned} \|\varphi_{\boldsymbol{\sigma}}(x_0) - \varphi_{\boldsymbol{\sigma}}(x_1)\| &\leq \delta \cdot (6/r) \cdot \rho(x_0, x_1) \\ &\leq C_\alpha \cdot r^{\alpha - 1} \cdot \rho(x_0, x_1) \leq C_\alpha \cdot \rho(x_0, x_1)^\alpha, \end{aligned}$$

since $\alpha \leq 1$ and $\rho(x_0, x_1) \leq r$.

On the other hand, if $x_1 \notin B_{r/2}(w_{j_0})$ then $\rho(x_0, x_1) \geq \rho(x_1, w_{j_0}) - \rho(x_0, w_{j_0}) \geq r/6$ whilst

$$\|\varphi_{\boldsymbol{\sigma}}(x_0) - \varphi_{\boldsymbol{\sigma}}(x_1)\| \leq 2 \cdot \delta \leq 2 \cdot \left((C_\alpha/12) \cdot r^\alpha\right) \leq C_\alpha \cdot \rho(x_0, x_1)^\alpha.$$

■

Given a smooth manifold $x_0, x_1 \in \mathcal{M}$ we let $\rho_g(x_0, x_1)$ denote the geodesic distance and for $r > 0$ we let $B_r^g(x_0)$ denote the geodesic metric ball of radius $r$. Given a set $A \subset \mathcal{M}$ and $r > 0$, an $r$-separated subset of $A$ is a set $\mathcal{W}_r = \{w_j\}_{j=1}^Q \subseteq A$ such that for $j_0 \neq j_1$, $\rho(w_{j_0}, w_{j_1}) \|w_{j_0}, w_{j_1}\|_2 > r$. A maximal $r$-separated subset is an $r$-separated subset of maximal cardinality. Note that $r$-separation is with respect to the Euclidean metric, rather than the geodesic metric.

**Lemma A.5** *Suppose $\mathcal{M} \subseteq \mathbb{R}^d$ is a compact smooth submanifold with dimension $\gamma$ and reach $\tau$. Fix $x_* \in \mathcal{M}$, $r_* > 0$. For each $r \in (0, r_*)$ we construct $S(r)$ by taking a maximal $r$-separated subset of $\overline{B^g_{r_*}(x_*)}$, $\mathcal{W}_r = \{w_j\}_{j=1}^{Q(r)}$, and defining*

$$S\left(r\right) := \bigcup_{j=1}^{Q(r)} \overline{B^g_{r/6}\left(w_j\right)},$$

*For all $r \in (0, r_*)$, $S\left(r\right)$ is a $\left(2^{-14\gamma}, \min\left\{r_*, \tau/8\right\}\right)$ regular set.*

To prove lemma A.5 we shall utilise the following geometric lemmas, the proof of which is given in appendix F. Let $v_\gamma$ denote the volume of the $\gamma$-dimensional Euclidean unit ball.

**Lemma A.6** *Let $\mathcal{M} \subseteq \mathbb{R}^d$ be a compact smooth submanifold with dimension $\gamma$, reach $\tau$ and Riemannian volume form $V_{\mathcal{M}}$. Then for all $x \in \mathcal{M}$ and $r < \tau/8$ we have*

$$4^{-\gamma} \cdot v_\gamma \cdot r^\gamma \leq V_{\mathcal{M}}\left(B_r^g(x)\right) \leq V_{\mathcal{M}}\left(B_r(x)\right) \leq 4^\gamma \cdot v_\gamma \cdot r^\gamma.$$

**Lemma A.7** *With the assumptions of lemma A.6, for all $x, \tilde{x} \in \mathcal{M}$ and $\tilde{r} \leq r < \tau/8$ with $\rho_g(x, \tilde{x}) \leq r + \tilde{r}/2$ we have $V_{\mathcal{M}}\left(B_r^g(x) \cap B_{\tilde{r}}^g(\tilde{x})\right) \geq 2^{-4\gamma} \cdot v_\gamma \cdot \tilde{r}^\gamma$.*

**Proof** [Proof of Lemma A.5] Fix $r \in (0, r_*)$ and take $\tilde{x} \in S(r)$ and $\tilde{r} \in (0, \min\{r_*, \tau/8\})$. We consider two cases.

Case 1: $\tilde{r} \geq 7r/3$. Let $\mathcal{J}(\tilde{x}, \tilde{r}) = \left\{j \in \{1, \cdots, Q(r)\} : \overline{B_r(w_j)} \cap B_{\tilde{r}/2}(\tilde{x}) \neq \emptyset\right\}$. Given $j \in \mathcal{J}(\tilde{x}, \tilde{r})$, we have $\|\tilde{x} - w_j\|_2 < r + \tilde{r}/2$. Thus, if $z \in B_{r/6}^g(w_j)$ then $\|z - \tilde{x}\|_2 \leq \rho_g(z, w_j) + \|\tilde{x} - w_j\|_2 < 7r/6 + \tilde{r}/2 \leq \tilde{r}$. Thus,

$$S(r) \cap B_{\tilde{r}}(\tilde{x}) \supseteq \bigcup_{j \in \mathcal{J}(\tilde{x}, \tilde{r})} B_{r/6}^g(w_j).$$

Since $\mathcal{W}_r$ is $r$-separated, for $j_0 \neq j_1$ we have $\rho_g(w_{j_0}, w_{j_1}) \geq \|w_{j_0} - w_{j_1}\|_2 \geq r$, so $B_{r/6}^g(w_{j_0}) \cap B_{r/6}^g(w_{j_1}) = \emptyset$. Hence, we may apply Lemma A.6 to obtain

$$V_{\mathcal{M}}\left(S(r) \cap B_{\tilde{r}}(\tilde{x})\right) \geq \sum_{j \in \mathcal{J}(\tilde{x}, \tilde{r})} V_{\mathcal{M}}\left(B_{r/6}^g(w_j)\right) \geq \#\mathcal{J}(\tilde{x}, \tilde{r}) \cdot 4^{-\gamma} \cdot v_\gamma \cdot (r/6)^\gamma.$$

Now we shall give a lower bound on $\#\mathcal{J}(\tilde{x}, \tilde{r})$. First note that since $\tilde{x} \in S(r)$ and $\mathcal{W}_r \subset \overline{B_{r_*}^g(x_*)}$ we have $\rho_g(x, \tilde{x}) \leq r_* + r/6 < r_* + \tilde{r}/4$. Moreover, $\tilde{r}/2 \in (0, \min\{r_*, \tau/8\})$, so by Lemma A.7 we have

$$V_{\mathcal{M}}\left(B_{r_*}^g(x_*) \cap B_{\tilde{r}/2}(\tilde{x})\right) \geq 2^{-5\gamma} \cdot v_\gamma \cdot \tilde{r}^\gamma.$$

By the maximality of $\mathcal{W}_r$ we have $B_{r_*}^g(x_*) \subseteq \bigcup_{j=1}^{Q(r)} \overline{B_r(w_j)}$, so $B_{r_*}^g(x_*) \cap B_{\tilde{r}/2}(\tilde{x}) \subseteq \bigcup_{j \in \mathcal{J}(\tilde{x}, \tilde{r})} \overline{B_r(w_j)}$. Hence,

$$2^{-5\gamma} \cdot v_\gamma \cdot \tilde{r}^\gamma \leq V_{\mathcal{M}}\left(B_{r_*}^g(x_*) \cap B_{\tilde{r}/2}(\tilde{x})\right) \leq \sum_{j \in \mathcal{J}(\tilde{x}, \tilde{r})} V_{\mathcal{M}}\left(\overline{B_r(w_j)}\right) \leq \#\mathcal{J}(\tilde{x}, \tilde{r}) \cdot 4^\gamma \cdot v_\gamma \cdot r^\gamma.$$

Thus, $\#\mathcal{J}(\tilde{x}, \tilde{r}) \geq 2^{-7\gamma} \cdot (r/\tilde{r})^\gamma$ and

$$V_{\mathcal{M}}\left(S(r) \cap B_{\tilde{r}}(\tilde{x})\right) \geq 2^{-12\gamma} \cdot v_\gamma \cdot \tilde{r}^\gamma \geq 2^{-14\gamma} \cdot V_{\mathcal{M}}\left(B_{\tilde{r}}(\tilde{x})\right).$$

Case 2: $\tilde{r} < 7r/3$. Since $\tilde{x} \in S(r)$ we may take $j_{\tilde{x}} \in \{1, \cdots, Q(r)\}$ so that $\tilde{x} \in \overline{B_{r/6}^g(w_{j_{\tilde{x}}})}$. By Lemma A.7 we have

$$V_{\mathcal{M}}\left(S(r) \cap B_{\tilde{r}}(\tilde{x})\right) \geq V_{\mathcal{M}}\left(B_{r/6}^g(w_{j_{\tilde{x}}}) \cap B_{\tilde{r}/14}(\tilde{x})\right) \geq 2^{-8\gamma} \cdot v_\gamma \cdot \tilde{r}^\gamma \geq 2^{-10\gamma} \cdot V_{\mathcal{M}}\left(B_{\tilde{r}}(\tilde{x})\right).$$

■

**Lemma A.8** *Fix $x_* \in \mathcal{M}$, let $\tilde{\tau} = \min\{\tau, 1\}$ and take $r_* = \tilde{\tau}/16$. Choose $r < r_*$ and let $S(r)$ and $Q(r)$ by as in the statement of Lemma A.5. Then $Q(r) \geq \left(2^{-8} \cdot \tilde{\tau}\right)^{\gamma} \cdot r^{-\gamma}$.*

**Proof** Note that $\mathcal{W}_r = \{w_j\}_{j=1}^{Q(r)}$ is a maximal $(\rho_g, r)$-separated subset of $\overline{B^g_{r_*}(x_*)}$. Hence, $B^g_{r_*}(x_*) \subseteq \bigcup_{j=1} \overline{B^g_r(w_j)}$. Thus,

$$2^{-6\gamma} \cdot v_\gamma \cdot \tilde{\tau}^\gamma \leq V_{\mathcal{M}}\left(B^g_{\tilde{\tau}/16}(x_*)\right) \leq \sum_{j=1}^{Q(r)} V_{\mathcal{M}}\left(\overline{B_r(w_j)}\right) \leq Q(r) \cdot 2^{2\gamma} \cdot v_\gamma \cdot r^\gamma.$$

■

**Lemma A.9** *Let $S(r)$ be as in the statement of Lemmas A.5 and A.8 with $r_* = \tilde{\tau}/16$. Then*

$$(3^{-1} \cdot 2^{-12} \cdot \tilde{\tau})^\gamma \cdot v_\gamma \leq V_{\mathcal{M}}(S(r)) \leq v_\gamma \cdot (\tilde{\tau}/2)^\gamma.$$

**Proof** Since $r \in (0, r_*)$ we must have $S(r) \subset B_{2r_*}(x_*)$, so it follows from Lemma A.6 that $V_{\mathcal{M}}(S(r)) \leq 4^\gamma \cdot v_\gamma \cdot (2r_*)^\gamma \leq v_\gamma \cdot (\tilde{\tau}/2)^\gamma$. On the other hand, since $\mathcal{W}_r$ is $r$-separated the balls $\left\{\overline{B^g_{r/6}(w_j)}\right\}_{j=1}^{Q(r)}$ are disjoint. Hence, combining Lemmas A.6 and A.8 we have

$$V_{\mathcal{M}}(S(r)) \geq \sum_{j=1}^{Q(r)} V_{\mathcal{M}}\left(B^g_{r/6}(w_j)\right) \geq Q(r) \cdot 3 \cdot 2^{-3\gamma} \cdot v_\gamma \cdot r^\gamma \geq (3^{-1} \cdot 2^{-11} \cdot \tilde{\tau})^\gamma \cdot v_\gamma.$$

■

We are now well placed to complete the proof of Proposition A.1.

**Proof** [Proof of Proposition A.1]

We take $\kappa_\Phi$ as in the statement of Lemma A.1 and define $Z_\Phi = M_\Phi(\boldsymbol{n}(\kappa_\Phi)) > 0$. Take $c_0 \in \left(0, 2^{-14\gamma}\right)$, $r_0 \in (0, \tilde{\tau}/16)$, $\nu_{\min} \in \left(0, (2/\tilde{\tau})^\gamma \cdot v_\gamma^{-1}\right)$, $\nu_{\max} \in \left((2^{14}/\tilde{\tau})^\gamma \cdot v_\gamma^{-1}, \infty\right)$, $\zeta_{\max} \in (0, Z_\Phi)$, $\alpha \in (0,1)$, $\beta \in (0, \gamma/\alpha)$, $C_\alpha, C_\beta > 0$ and let

$$\Gamma = \langle (r_0, c_0, \nu_{\min}, \nu_{\max}), (\beta, C_\beta, \zeta_{\max}), (\alpha, C_\alpha) \rangle.$$

For each $r \in (0, r_0)$ we shall construct an associated set of probability measures $\mathcal{P}(r) \subset \mathcal{P}(\mathcal{M}, \Gamma)$ as follows. Fix some $r \in (0, r_0)$. We begin by constructing $\mu$, which will be common to all $\mathbb{P} \in \mathcal{P}(r)$. To do so we take $S(r)$ as in the statement of Lemma A.5. That is, we fix some $x_* \in \mathcal{M}$ and construct $S(r)$ by taking $\mathcal{W}_r = \{w_j\}_{j=1}^{Q(r)}$ to be a maximal $r$-separated subset of $B^g_{r_*}(x_*)$ and define,

$$S(r) = \bigcup_{j=1}^{Q(r)} \overline{B^g_{r/6}(w_j)}.$$

Let $\nu_*(r) = (V_{\mathcal{M}}(S(r)))^{-1}$. We let $\mu$ be a probability measure which is absolutely continuous with respect to $V_{\mathcal{M}}$ and has density

$$\nu(x) = \begin{cases} \nu_*(r) & \text{if } x \in S(r) \\ 0 & \text{if } x \notin S(r). \end{cases}$$

Clearly, $\text{supp}(\mu) = S(r)$ and by Lemma A.5 the set $S(r)$ is $(c_0, r_0)$ regular. Moreover, by Lemma A.9 we have

$$\nu_{\min} < (2/\tilde{\tau})^{\gamma} \cdot v_{\gamma}^{-1} \leq \nu_*(r) \leq (2^{14}/\tilde{\tau})^{\gamma} \cdot v_{\gamma}^{-1} < \nu_{\max}.$$

Hence, the measure $\mu$ is $(c_0, r_0, \nu_{\min}, \nu_{\max})$ regular. For each $j = \{1, \cdots, Q(r)\}$ we let $B_j = B^g_{r/6}(w_j)$ and $A_j = B_{r/3}(w_j)$. Since $\mathcal{W}_r$ is $r$-separated, the balls $A_j$ are disjoint. Hence, $\mu(A_j \backslash B_j) = 0$, since $\mu$ is absolutely continuous and supported on $S(r)$. In addition, by Lemma A.6 we have $v(r) \leq \mu(B_j) \leq \mu(A_j) \leq u(r)$ with $v(r) = \nu_*(r) \cdot 4^{-\gamma} \cdot v_{\gamma} \cdot (r/6)^{\gamma}$ and $u(r) = \nu_*(r) \cdot 4^{\gamma} \cdot \nu_{\gamma} \cdot (r/6)^{\gamma}$. Take $t_{\Phi}$ as in the statement of Lemma A.1 and let $\delta(r) = \min\{t_{\Phi}/2, (C_{\alpha}/12) \cdot r^{\alpha}\}$. In addition, we take $m(r) = \min\{Q(r), \lfloor (C_{\beta} \cdot c_{\Phi})^{\beta} \cdot u(r)^{-1} \cdot \delta(r)^{\beta} \rfloor\}$. By Lemma A.8 we have $Q(r) \geq \left(2^{-8} \cdot \tilde{\tau}\right)^{\gamma} \cdot r^{-\gamma}$. Hence, there exists constants $R(\Gamma), m_0(\Gamma) > 0$ such that for all $r < R(\Gamma)$, $\delta(r) = (C_{\alpha}/12) \cdot r^{\alpha}$ and $m(r) \geq m_0(\Gamma) \cdot r^{\alpha \cdot \beta - \gamma}$. By Lemma A.4 there exists functions $\{g_j\}_{j=1}^m$ so that $g_j(x) = 1$ for all $x \in B_j$, $g_j(x) = 0$ for all $x \notin A_j$ and $\boldsymbol{\sigma} \in \{-1, 0, +1\}^m$, the function $\eta_{\boldsymbol{\sigma}} : \mathcal{X} \to \mathbb{R}^{L \times 1}$ by

$$\eta_{\boldsymbol{\sigma}}(x) = \boldsymbol{n}\left(\kappa_{\Phi} + \delta \cdot \sum_{j=1}^{m} \sigma_j \cdot g_j(x)\right),$$

is Hölder continuous with constants $(\alpha, C_{\alpha})$. For each $\boldsymbol{\sigma} \in \{-1, 0, +1\}^m$ we let $\mathbb{P}_{\boldsymbol{\sigma}}$ denote the measure on $\mathcal{Z} = \mathcal{X} \times \mathcal{Y}$ formed by taking $\mu$ to be the marginal distribution over $\mathcal{X}$ and $\eta_{\boldsymbol{\sigma}}$ to be the conditional distribution of $\mathcal{Y}$ given $x \in \mathcal{X}$. Since $\delta(r) \in (0, t_{\Phi})$ and $m(r) \cdot u(r) \leq C_{\beta} \cdot (c_{\Phi} \cdot \delta(r))^{\beta}$ and $\zeta_{\max} \in (0, Z_{\Phi})$, it follows from Lemma A.3 that each $\mathbb{P}_{\boldsymbol{\sigma}}$ satisfies the margin condition with constants $(\beta, C_{\beta}, \zeta_{\max})$. We let

$$\mathcal{P}(r) := \{\mathbb{P}_{\boldsymbol{\sigma}} : \boldsymbol{\sigma} \in \{-1, +1\}^m\}.$$

We have shown that $\mu$ is $(c_0, r_0, \nu_{\min}, \nu_{\max})$ regular each $\eta_{\boldsymbol{\sigma}}$ is Hölder continuous with constants $(\alpha, C_{\alpha})$ and each $\mathbb{P}_{\boldsymbol{\sigma}}$ satisfies the margin condition with constants $(\beta, C_{\beta}, \zeta_{\max})$. Thus, $\mathcal{P}(r) \subset \mathbb{P} \in \mathcal{P}_{\Phi}(V_{\mathcal{M}}, \Gamma)$. Hence, by Lemma A.2, for all classifiers $\hat{f}$ and all $n \in \mathbb{N}$ we have

$$\sup_{\mathbb{P} \in \mathcal{P}_{\Phi}(V_{\mathcal{M}}, \Gamma)} \left\{\mathbb{E}_{\mathbb{P}^n}\left[R_{\mathbb{P}}(\hat{f}_n)\right] - R_{\mathbb{P}}(f^*_{\mathbb{P}})\right\}$$

$$\geq (2c_{\Phi} \cdot m(r) \cdot v(r)) \cdot \delta(r) \cdot \left(1 - 2\delta(r) \cdot \sqrt{\frac{n \cdot u(r)}{t_{\Phi}}}\right).$$

It follows from the construction of $\delta(r)$, $u(r)$, $v(r)$, $m(r)$ and Lemma A.9 we see that there exists $R', C', C'' > 0$, depending purely upon $\Phi$, $\gamma$, $\tau$ and $\Gamma$, such that for all $r < R'$,

$$\delta(r)^2 \cdot u(r) \cdot t_{\Phi}^{-1} < C' \cdot r^{\gamma + 2\alpha}$$

$$c_{\Phi} \cdot m(r) \cdot v(r) \cdot \delta(r) > C'' \cdot r^{(\alpha\beta - \gamma) + \gamma + \alpha} = C'' \cdot r^{\alpha(\beta+1)}.$$

Thus, if we take $r = \min\{R', (16C')^{-1}\} \cdot n^{-\frac{1}{2\alpha+\gamma}}$ we have $2\delta(r) \cdot \sqrt{n \cdot u(r) \cdot t_{\Phi}}^{-1} < 1/2$. Hence,

$$\sup_{\mathbb{P} \in \mathcal{P}_{\Phi}(V_{\mathcal{M}}, \Gamma)} \left\{\mathbb{E}_{\mathbb{P}^n}\left[R_{\mathbb{P}}(\hat{f}_n)\right] - R_{\mathbb{P}}(f^*_{\mathbb{P}})\right\} \geq C'' \cdot \left(\min\{R', (16C')^{-1}\}\right)^{\alpha(\beta+1)} \cdot n^{-\frac{\alpha(\beta+1)}{2\alpha+\gamma}}.$$

This completes the proof of Proposition A.1. ■

## Appendix B. Bounding the probability of mis-classification

In this section we prove Proposition B.1 which forms the first part of Theorem 5.

**Proposition B.1** *Suppose that $\mathbb{P}$ satisfies the margin condition with constants $(\beta, C_\beta, \zeta_{\max})$ and that the conditional $\eta$ is measure-smooth, with constants $(\lambda, C_\lambda)$. Suppose that generates $\omega$ measure-approximate nearest neighbours with respect to the measure $\mu$. Take $k_0 > 0$ and for each $n \in \mathbb{N}$ take $k_n = k_0 \cdot n^{\frac{2\lambda}{2\lambda+1}} \cdot (1+\log(1/\delta))^{1/(2\lambda+1)}$. There exists a constant $C > 0$ depending purely upon $C_\lambda, \lambda, C_\beta, \beta, k_0$ such that for all $n \in \mathbb{N}$, with probability at least $1-\delta$ over $\mathcal{D}_n \sim \mathbb{P}^n$ we have*

$$\mathbb{P}\left[f^S_{n,k}(X) \notin \mathcal{Y}^*_\Phi\left(\eta(X)\right)\right] \leq \delta + C \cdot \left(\omega^\lambda \cdot \Lambda(\Phi) \cdot \sqrt{\log(L)}\right)^\beta \cdot \left(\frac{1+\log(1/\delta)}{n}\right)^{\beta\lambda/(2\lambda+1)}.$$

To prove Proposition B.1 we first introduce some notation before giving some preliminary lemmas. We define $r_p(x) = \inf\left\{r > 0 : \mu\left(B_r(x)\right) \geq p\right\}$. We make use of the following standard results.

**Lemma B.1** *Take $x \in \mathcal{X}$. Suppose that $p \in [0,1]$, $\xi \leq 1$, $k \leq (1-\xi)np$. Then*

$$\mathbb{P}^n\left[\max\left\{\rho(x, X_i) : i \in S^\circ_k(x, \mathcal{F}_n)\right\} > r_p(x)\right] \leq \exp(-k\xi^2/2).$$

**Lemma B.2** *Take $x \in \mathcal{X}$. For each $\delta > 0$ we have*

$$\mathbb{P}^n\left[\left\|\hat{\eta}^S_{n,k}(x) - \frac{1}{k}\sum_{i \in S_k(x,\mathcal{F}_n)} \eta(X_i)\right\|_\infty \geq \delta | \mathcal{F}_n\right] \leq 2L\exp(-2k\delta^2).$$

The proof of lemmas B.1 and B.2 is given in Appendix E. Recall that we defined $\Lambda(\Phi)$ by

$$\Lambda(\Phi) := (L-2)\cdot \mathrm{Asym}(\Phi) + 2\|\Phi\|_\infty.$$

**Lemma B.3** *Given $y_0, y_1 \in \mathcal{Y}$ and $\boldsymbol{n}_0, \boldsymbol{n}_1 \in \Delta(\mathcal{Y})$ we have*

$$\|\left(e(y_1) - e(y_0)\right)^T \ \Phi\left(\boldsymbol{n}_0 - \boldsymbol{n}_1\right)\| \leq \Lambda(\Phi) \cdot \|\boldsymbol{n}_0 - \boldsymbol{n}_1\|_\infty.$$

**Proof** This follows immediately from the definitions. ∎

Given $p \in (0,1)$, and $\Delta > 0$ we define

$$\mathcal{X}_{p,\Delta} = \Big\{x \in \mathcal{X} : \forall \tilde{x} \in \overline{B_{r_p(x)}(x)},\ y_0 \in \mathcal{Y}^*_\Phi\left(\eta(x)\right),\ y_1 \in \mathcal{Y}\backslash\mathcal{Y}^*_\Phi\left(\eta(x)\right), \\ \left(e(y_1) - e(y_0)\right)^T \Phi\, \eta(\tilde{x}) \geq \Lambda(\Phi)\cdot\Delta\Big\},$$

and let $\partial^\theta_{p,\Delta} = \mathcal{X}\backslash\mathcal{X}^\theta_{p,\Delta}$.

**Lemma B.4** *Suppose that $k < n$ and $S$ generates $\omega$ measure approximate nearest neighbours. Take $p \in (0,1)$ and $\Delta > 0$ and suppose that $x \in \mathcal{X}_{p,\Delta}$ satisfies both*

*1.* $\max\{\rho(x,X_i) : i \in S_k^{\circ}(x,\mathcal{F}_n)\} \leq r_{p/\omega}(x)$

*2.* $\left\|\left|\hat{\eta}_{n,k}^S(x) - \frac{1}{k}\sum_{i\in S_k(x,\mathcal{F}_n)} \eta(X_i)\right|\right\|_\infty < \Delta.$

*Then* $f_{n,k_n}^S(x) \in \mathcal{Y}_\Phi^* (\eta(x))$.

**Proof** Since $\max\{\rho(x,X_i) : i \in S_k^{\circ}(x,\mathcal{F}_n)\} \leq r_{p/\omega}(x)$ and $S$ generates $\omega$ measure-approximate nearest neighbours we have $\max\{\rho(x,X_i) : i \in S_k(x,\mathcal{F}_n)\} \leq r_p(x)$. Since $x \in \mathcal{X}_{p,\Delta}^{\theta}$ for any $y_0 \in \mathcal{Y}_\Phi^* (\eta(x))$, $y_1 \in \mathcal{Y}\backslash\mathcal{Y}_\Phi^* (\eta(x))$ we have $(e(y_1)-e(y_0))^T \Phi\, \eta(X_i) \geq \Lambda(\Phi)\cdot\Delta$ for all $i \in S_k(x,\mathcal{F}_n)$, so

$$(e(y_1)-e(y_0))^T\, \Phi \left(\frac{1}{k}\sum_{i\in S_k(x,\mathcal{F}_n)} \eta(X_i)\right) \geq \Lambda(\Phi)\cdot\Delta.$$

Moreover, since $\left\|\left|\hat{\eta}_{n,k}^S(x) - \frac{1}{k}\sum_{i\in S_k(x,\mathcal{F}_n)} \eta(X_i)\right|\right\|_\infty < \Delta$, by Lemma B.3, this implies that for all $y_0 \in \mathcal{Y}_\Phi^* (\eta(x))$ and $y_1 \in \mathcal{Y}\backslash\mathcal{Y}_\Phi^* (\eta(x))$ we have

$$(e(y_0))^T\, \Phi\, \hat{\eta}_{n,k}^S(x) < (e(y_1))^T\, \Phi\, \hat{\eta}_{n,k}^S(x).$$

Thus, $\mathcal{Y}_\Phi^* (\eta(x)) = \mathcal{Y}_\Phi^* \left(\hat{\eta}_{n,k}^S(x)\right)$. Hence, $f_{n,k_n}^S(x) \in \mathcal{Y}_\Phi^* (\eta(x))$. ∎

**Lemma B.5** *Take $\delta \in (0,1)$, $k\cdot\omega < n$ and suppose that $S$ generates $\omega$ measure approximate $k$-nearest neighbours. With probability at least $1-\delta$ over $\mathcal{D}_n \sim \mathbb{P}^n$ we have*

$$\mathbb{P}\left[f_{n,k}^S(X) \notin \mathcal{Y}_\Phi^* (\eta(X))\right] \leq \delta + \mu\left(\partial_{p,\Delta}\right),$$

*where*

$$p = \frac{k}{n}\cdot\frac{\omega}{1-\sqrt{(2/k)\log(2/\delta^2)}} \qquad \textit{and} \qquad \Delta = \sqrt{\frac{1}{2k}\log\frac{4L}{\delta^2}}.$$

**Proof** Given $\mathcal{D}_n \sim \mathbb{P}^n$ we define $A(\mathcal{D}_n) = \left\{x \in \mathcal{X}_{p,\Delta} : f_{n,k}^S(X) \notin \mathcal{Y}_\Phi^* (\eta(X))\right\}$. To prove the lemma it suffices to show that with probability at least $1-\delta$ over $\mathcal{D}_n \sim \mathbb{P}^n$ we have $\mathbb{P}[A(\mathcal{D}_n)] \leq \delta$. This follows from the definition of $\partial_{p,\Delta}$ as $\mathcal{X}\backslash\mathcal{X}_{p,\Delta}$. Now by Lemma B.4 we have

$$\begin{aligned} A(\mathcal{D}_n) \subseteq &\left\{x \in \mathcal{X} : \max\{\rho(x,X_i) : i \in S_k^{\circ}(x,\mathcal{F}_n)\} > r_{p/\omega}(x)\right\} \\ &\cup \left\{x \in \mathcal{X} : \left\|\left|\hat{\eta}_{n,k}^S(x) - \frac{1}{k}\sum_{i\in S_k(x,\mathcal{F}_n)} \eta(X_i)\right|\right\|_\infty \geq \Delta\right\}. \end{aligned}$$

Now take $x \in \mathcal{X}$. By Lemma B.1 with $\xi = 1-(k\omega)/(np)$ we have

$$\mathbb{P}^n\left[\max\{\rho(x,X_i) : i \in S_k^{\circ}(x,\mathcal{F}_n)\} > r_{p/\omega}(x)\right] \leq \exp(-k\xi^2/2) = \frac{\delta^2}{2}.$$

By Lemma B.2 we have

$$\mathbb{P}^n\left[\left\|\hat{\eta}^S_{n,k}(x) - \frac{1}{k}\sum_{i \in S_k(x,\mathcal{F}_n)} \eta(X_i)\right\|_\infty \geq \Delta\right] \leq 2L\exp(-2k\Delta^2) = \frac{\delta^2}{2}.$$

Hence, $\mathbb{P}^n\left[x \in A(\mathcal{D}_n)\right] \leq \delta^2$. Since this holds for all $x \in \mathcal{X}$, by Fubini's theorem we have

$$\begin{aligned}\mathbb{E}_n\left[\mathbb{P}\left[A(\mathcal{D}_n)\right]\right] &= \mathbb{E}_n\left[\mathbb{E}\left[\mathbb{1}_{X \in A(\mathcal{D}_n)}\right]\right] \\ &= \mathbb{E}\left[\mathbb{E}_n\left[\mathbb{1}_{X \in A(\mathcal{D}_n)}\right]\right] = \mathbb{E}\left[\mathbb{P}^n\left[X \in A(\mathcal{D}_n)\right]\right] \leq \delta^2.\end{aligned}$$

Thus, by Markov's inequality $\mathbb{P}^n\left[\mathbb{P}\left[A(\mathcal{D}_n)\right] > \delta\right] \leq \delta$. Thus, with probability at least $1-\delta$ over $\mathcal{D}_n \sim \mathbb{P}^n$ we have $\mathbb{P}\left[A(\mathcal{D}_n)\right] \leq \delta$ and the lemma holds. ■

**Lemma B.6** *Suppose that $\mathbb{P}$ satisfies the margin condition with constants $(\beta, C_\beta, \zeta_{\max})$ and that the conditional $\eta$ is measure-smooth, with constants $(\lambda, C_\lambda)$. Given any $\Delta > 0$ and $p \in (0,1)$ satisfying $\Lambda(\Phi) \cdot \left(\Delta + C_\lambda \cdot p^\lambda\right) < \zeta_{\max}$ we have*

$$\mu\left(\partial_{p,\Delta}\right) \leq C_\beta \cdot \left(\Lambda(\Phi) \cdot \left(\Delta + C_\lambda \cdot p^\lambda\right)\right)^\beta.$$

**Proof** Suppose that $x \in \partial_{p,\Delta}$. Then there exists some $\tilde{x} \in \mathcal{X}$ with $\rho(\tilde{x}, x) \leq r_p(x)$, some $y_0 \in \mathcal{Y}^*_\Phi\left(\eta(x)\right)$ and some $y_1 \in \mathcal{Y} \backslash \mathcal{Y}^*_\Phi\left(\eta(x)\right)$ such that $\left(e(y_1) - e(y_0)\right)^T \Phi\, \eta(\tilde{x}) < \Lambda(\Phi) \cdot \Delta$. Since $\eta$ is measure smooth with constants $(\lambda, C_\lambda)$ we have,

$$\|\eta(\tilde{x}) - \eta(x)\|_\infty \leq C_\lambda \cdot \mu\left(B_{\rho(\tilde{x},x)}(x)\right)^\lambda \leq C_\lambda \cdot \mu\left(B_{r_p(x)}(x)\right)^\lambda \leq C_\lambda \cdot p^\lambda.$$

By Lemma B.3 this implies that

$$M_\Phi\left(\eta(x)\right) = \left(e(y_1) - e(y_0)\right)^T \Phi\, \eta(x) < \Lambda(\Phi) \cdot \left(\Delta + C_\lambda \cdot p^\lambda\right).$$

Hence,

$$\begin{aligned}\mu\left(\partial_{p,\Delta}\right) &\leq \mu\left(\left\{x \in \mathcal{X} : M_\Phi(\eta(x)) < \Lambda(\Phi) \cdot \left(\Delta + C_\lambda \cdot p^\lambda\right)\right\}\right) \\ &\leq C_\beta \cdot \left(\Lambda(\Phi) \cdot \left(\Delta + C_\lambda \cdot p^\lambda\right)\right)^\beta.\end{aligned}$$

■

**Proof** [Proof of Proposition B.1] First note that without loss of generality we may assume that $\zeta_{\max} = \infty$. Indeed if $\mathbb{P}$ satisfies the margin condition with constants $(\beta, C_\beta, \zeta_{\max})$ then $\mathbb{P}$ also satisfies the margin condition with constants $(\tilde{C}_\beta, \beta, \infty)$, where $\tilde{C}_\beta = \max\{C_\beta, 2\zeta_{\max}^{-\beta}\}$.

To complete the proof, for each $n \in \mathbb{N}$, we take

$$\begin{aligned}k_n &= k_0 \cdot n^{\frac{2\lambda}{2\lambda+1}} \left(1 + \log(1/\delta)\right)^{1/(2\lambda+1)} \\ p_n &= (k_n \omega) / \left(n\left(1 - \sqrt{(2/k_n)\log(2/\delta^2)}\right)\right) \\ \Delta_n &= \sqrt{(1/2k_n)\log\left(4L/\delta^2\right)},\end{aligned}$$

and apply lemmas B.5 and B.6. Indeed, suppose that $n \geq (16/k_0)^{(2\lambda+1)/(2\lambda)}(1+\log(1/\delta))$. It follows that $\sqrt{(2/k_n)\log(2/\delta^2)} < 1/2$, so

$$p_n < (2\omega)\cdot\frac{k_n}{n} = (2k_0\omega)\cdot\left(\frac{1+\log(1/\delta)}{n}\right)^{1/(2\lambda+1)}.$$

In addition, for some constant $c_{k_0}$, depending upon $k_0$, we have

$$\Delta_n \leq c_{k_0}\cdot\sqrt{\log(L)}\cdot\left(\frac{1+\log(1/\delta)}{n}\right)^{\lambda/(2\lambda+1)}.$$

Moreover, by lemmas B.5 and B.6, we have

$$\begin{aligned}\mathbb{P}\left[f^S_{n,k}(X)\notin\mathcal{Y}^*_\Phi\left(\eta(X)\right)\right] &\leq \delta+\mu\left(\partial_{p,\Delta}\right)\\ &\leq \delta + C_\beta\cdot\left(\Lambda(\Phi)\cdot\left(\Delta_n+C_\lambda\cdot p_n^\lambda\right)\right)^\beta\\ &\leq C\cdot\left(\omega^\lambda\cdot\Lambda(\Phi)\cdot\sqrt{\log(L)}\right)^\beta\cdot\left(\frac{1+\log(1/\delta)}{n}\right)^{\beta\lambda/(2\lambda+1)},\end{aligned}$$

where $C$ is a constant depending purely upon $k_0, \beta, \lambda, \zeta_{\max}, C_\beta, C_\lambda$. By increasing the constant, depending upon $k_0, \beta, \lambda$, the bound also holds for $n < (16/k_0)^{(2\lambda+1)/(2\lambda)}(1+\log(1/\delta))$. ■

## Appendix C. Bounding the expected risk

In this section we prove Proposition C.1 which forms the second part of Theorem 5.

**Proposition C.1** *Suppose that $\mathbb{P}$ satisfies the margin condition with constants $(C_\beta, \beta, \zeta_{\max})$ and that the conditional $\eta$ is measure-smooth, with constants $(\lambda, C_\lambda)$. Suppose that generates $\omega$ measure-approximate nearest neighbours with respect to the measure $\mu$. Take $k_0 > 0$ and for each $n \in \mathbb{N}$ take $k_n = k_0 \cdot n^{\frac{2\lambda}{2\lambda+1}}$. There exists a constant $C > 0$ depending purely upon $C_\lambda, \lambda, C_\beta, \beta, k_0$ such that for all $n \in \mathbb{N}$ we have*

$$\mathbb{E}_n\left[R\left(f^S_{n,k_n}\right)\right] - R^* \leq C\cdot L\cdot\left(\omega^\lambda\cdot\Lambda(\Phi)\right)^{1+\beta}\cdot n^{-\frac{\lambda(1+\beta)}{2\lambda+1}}.$$

Proposition C.1 follows from Propositions C.2 and C.3.

**Proposition C.2** *Take a probability distribution $\mathbb{P}$ on $\mathcal{Z} = \mathcal{X}\times\mathcal{Y}$ determined by a marginal distribution $\mu$ and conditional probability $\eta$. Suppose that for each $n \in \mathbb{N}$, $\hat{\eta}_n : \mathcal{X} \to \mathbb{R}^L$ is an estimator of $\eta$ determined by $\mathcal{D}_n \sim \mathbb{P}^n$. Suppose further that there exists constants $C_1, C_2 > 0$, $N_0 \in \mathbb{N}$ and some decreasing positive sequence $(a_n)_{n\geq N_0}$ such that for each $n \geq N_0$, $\mu$ almost every $x \in \mathcal{X}$ there exists a set $A_n(x) \subseteq \mathcal{Z}^n$ such that $\mathbb{P}^n\left[A_n(x)\right] \leq a_n^{1+\beta}$ and for all $\xi \geq a_n$ we have*

$$\mathbb{P}^n\left[\|\hat{\eta}_n(x)-\eta(x)\|_\infty > \xi | \mathcal{D}_n \notin A_n(x)\right] \leq C_1\cdot\exp\left(-C_2\cdot\left(\frac{\xi}{a_n}\right)^2\right).$$

*Suppose for each $n \in \mathbb{N}$ we construct a classifier $\hat{f}_n : \mathcal{X} \to \mathcal{Y}$, based on $\hat{\eta}_n : \mathcal{X} \to \mathbb{R}^L$ and defined by $\hat{f}_n(x) = \min\left(\mathcal{Y}^*_\Phi\left(\hat{\eta}_n(x)\right)\right)$. If $\mathbb{P}$ satisfies the margin condition with constants $(C_\beta, \beta, \zeta_{\max})$ then there exists $C = C\left(C_2, C_\beta, \beta, \zeta_{\max}\right) > 0$, which is monotonically increasing with $\beta$, such that for all $n \geq N_0$ we have*

$$\mathbb{E}_n\left[R\left(\hat{f}_n\right)\right] - R^* \leq C \cdot (1 + C_1) \cdot (\Lambda(\Phi) \cdot a_n)^{1+\beta}.$$

**Proposition C.3** *Suppose that the conditional $\eta$ is measure-smooth, with constants $(\lambda, C_\lambda)$, and that $S$ generates $\omega$ measure-approximate nearest neighbours. Then for any $n \in \mathbb{N}$, $k \leq n/2$ if for $\mu$ almost every $x \in \mathcal{X}$ we let*

$$A_n(x) = \left\{\mathcal{D}_n : \max\left\{\rho(x, X_i) : i \in S^\circ_k(x, \mathcal{F}_n)\right\} > r_{2k/n}(x)\right\}.$$

*Then $\mathbb{P}^n\left[A_n(x)\right] \leq \exp(-k/8)$ and for all $\xi \geq 2C_\lambda \cdot \left(\frac{2\omega k}{n}\right)^\lambda$ we have*

$$\mathbb{P}^n\left[\left|\left|\hat{\eta}^S_{n,k}(x) - \eta(x)\right|\right|_\infty \geq \xi | \mathcal{D}_n \notin A_n(x)\right] \leq 2L\exp(-\frac{k\xi^2}{2}).$$

**Proof** [Proof of Proposition C.1] We combine Proposition C.2 with Proposition C.3, with $a_n = 2C_\lambda \cdot (2\omega k_0)^\lambda \cdot n^{-\frac{\lambda}{2\lambda+1}}$. ■

**Proof** [Proof of Proposition C.2] First note that as in the proof of Propostion B.1 we may assume that $\zeta_{\max} = \infty$, without loss of generality. We define a Bayes optimal classifier $f^* : \mathcal{X} \to \mathcal{Y}$ by $f^*(x) = \min\left(\mathcal{Y}^*_\Phi\left(\eta(x)\right)\right)$, so $R(f^*) = R^*$. Hence,

$$\begin{aligned}
&\mathbb{E}_n\left[R\left(\hat{f}_n\right)\right] - R^* \\
&= \mathbb{E}_n\left[R\left(\hat{f}_n\right) - R(f^*)\right] \\
&= \mathbb{E}_n\left[\mathbb{E}\left[\phi_{\hat{f}_n(X),Y} - \phi_{f^*(X),Y}\right]\right] \\
&= \mathbb{E}_n\left[\int \left(e(\hat{f}_n(x)) - e(f^*(x))\right)^T \Phi\, \eta(x) d\mu(x)\right] \\
&= \int \mathbb{E}_n\left[\left(e(\hat{f}_n(x)) - e(f^*(x))\right)^T \Phi\, \eta(x)\right] d\mu(x) \\
&\leq \int \left(\mathbb{E}_n\left[\left(e(\hat{f}_n(x)) - e(f^*(x))\right)^T \Phi\, \eta(x) | \mathcal{D}_n \notin A_n(x)\right] + 2 \cdot \|\Phi\|_\infty \cdot \mathbb{P}^n\left[A_n(x)\right]\right) d\mu(x) \\
&\leq \int \mathbb{E}_n\left[\left(e(\hat{f}_n(x)) - e(f^*(x))\right)^T \Phi\, \eta(x) | \mathcal{D}_n \notin A_n(x)\right] d\mu(x) + \Lambda(\Phi) \cdot a_n^{1+\beta}.
\end{aligned}$$

Thus, it suffices to show that there exists $\tilde{C} = \tilde{C}\left(C_2, C_\beta, \beta\right) > 0$, which is monotonically increasing with $\beta$, such that for all $n \geq N_0$,

$$\int \mathbb{E}_n\left[\left(e(\hat{f}_n(x)) - e(f^*(x))\right)^T \Phi\, \eta(x) | \mathcal{D}_n \notin A_n(x)\right] d\mu(x) \leq \tilde{C} \cdot (1 + C_1) \cdot (\Lambda(\Phi) \cdot a_n)^{1+\beta}. \tag{3}$$

We define sets $\Omega_j(\mathcal{D}_n)$ for each $j \in \mathbb{N}$ and $\mathcal{D}_n \sim \mathbb{P}^n$ by

$$\Omega_0(\mathcal{D}_n) = \left\{x \in \mathcal{X} : 0 < \left(e(\hat{f}_n(x)) - e(f^*(x))\right)^T \Phi\, \eta(x) < \Lambda(\Phi) \cdot a_n\right\}$$
$$\Omega_j(\mathcal{D}_n) = \left\{x \in \mathcal{X} : 2^{j-1} \cdot \Lambda(\Phi) \cdot a_n < \left(e(\hat{f}_n(x)) - e(f^*(x))\right)^T \Phi\, \eta(x) < 2^j \cdot \Lambda(\Phi) \cdot a_n\right\}.$$

Note that for all $j \geq 1$, if $x \in \Omega_j(\mathcal{D}_n)$ then

$$2^{j-1} \cdot \Lambda(\Phi) \cdot a_n < \left(e(\hat{f}_n(x)) - e(f^*(x))\right)^T \Phi\, \eta(x).$$

On the other hand, by the construction of $\hat{f}_n$ we have

$$\left(e(\hat{f}_n(x)) - e(f^*(x))\right)^T \Phi\, \hat{\eta}_n(x) \leq 0.$$

Thus, by Lemma B.3 for all $j \geq 1$, if $x \in \Omega_j(\mathcal{D}_n)$ then $\|\hat{\eta}_n(x) - \eta(x)\|_\infty > 2^{j-1} a_n \geq a_n$. Moreover, for all $j \in \mathbb{N}$, if $x \in \Omega_j(\mathcal{D}_n)$ then $M_\Phi\left(\eta(x)\right) < 2^j \cdot \Lambda(\Phi) \cdot a_n$. So if $j \geq 1$ we have

$$\mathbb{1}_{\Omega_j(\mathcal{D}_n)} \leq \mathbb{1}_{\{\|\hat{\eta}_n(x)-\eta(x)\|_\infty > 2^{j-1}a_n\}} \cdot \mathbb{1}_{\{M_\Phi(\eta(x)) < 2^j \cdot \Lambda(\Phi) \cdot a_n\}}.$$

Thus, $\int \mathbb{E}_n\left[\mathbb{1}_{\Omega_j(\mathcal{D}_n)} | \mathcal{D}_n \notin A_n(x)\right] d\mu(x)$

$$\begin{aligned}
&\leq \int \mathbb{E}_n\left[\mathbb{1}_{\{\|\hat{\eta}_n(x)-\eta(x)\|_\infty > 2^{j-1}a_n\}} \cdot \mathbb{1}_{\{M_\Phi(\eta(x)) < 2^j \cdot \Lambda(\Phi) \cdot a_n\}} | \mathcal{D}_n \notin A_n(x)\right] d\mu(x) \\
&= \int \mathbb{E}_n\left[\mathbb{1}_{\{\|\hat{\eta}_n(x)-\eta(x)\|_\infty > 2^{j-1}a_n\}} | \mathcal{D}_n \notin A_n(x)\right] \cdot \mathbb{1}_{\{M_\Phi(\eta(x)) < 2^j \cdot \Lambda(\Phi) \cdot a_n\}} d\mu(x) \\
&= \int \mathbb{P}^n\left[\|\hat{\eta}_n(x) - \eta(x)\|_\infty > 2^{j-1} a_n | \mathcal{D}_n \notin A_n(x)\right] \cdot \mathbb{1}_{\{M_\Phi(\eta(x)) < 2^j \cdot \Lambda(\Phi) \cdot a_n\}} d\mu(x) \\
&\leq \int C_1 \cdot \exp\left(-C_2 \cdot 4^{j-1}\right) \cdot \mathbb{1}_{\{M_\Phi(\eta(x)) < 2^j \cdot \Lambda(\Phi) \cdot a_n\}} d\mu(x) \\
&\leq C_1 \cdot \exp\left(-C_2 \cdot 4^{j-1}\right) \cdot \mu\left(\{M_\Phi\left(\eta(x)\right) < 2^j \cdot \Lambda(\Phi) \cdot a_n\}\right) \\
&\leq C_1 \cdot \exp\left(-C_2 \cdot 4^{j-1}\right) \cdot C_\beta \cdot \left(2^j \cdot \Lambda(\Phi) \cdot a_n\right)^\beta.
\end{aligned}$$

In addition $\int \mathbb{E}_n \left[\mathbb{1}_{\Omega_0(\mathcal{D}_n)} | \mathcal{D}_n \notin A_n(x)\right] d\mu(x) \leq C_\beta \cdot (\Lambda(\Phi) \cdot a_n)^\beta$. Thus,

$$\begin{aligned}
&\int \mathbb{E}_n \left[ \left( e(\hat{f}_n(x)) - e(f^*(x)) \right)^T \Phi \, \eta(x) | \mathcal{D}_n \notin A_n(x) \right] d\mu(x) \\
&= \int \mathbb{E}_n \left[ \sum_{j=0}^{\infty} \mathbb{1}_{\Omega_j(\mathcal{D}_n)} \left( e(\hat{f}_n(x)) - e(f^*(x)) \right)^T \Phi \, \eta(x) | \mathcal{D}_n \notin A_n(x) \right] d\mu(x) \\
&\leq (\Lambda(\Phi) \cdot a_n) \cdot \sum_{j=0}^{\infty} 2^j \cdot \int \mathbb{E}_n \left[ \mathbb{1}_{\Omega_j(\mathcal{D}_n)} | \mathcal{D}_n \notin A_n(x) \right] d\mu(x) \\
&\leq (\Lambda(\Phi) \cdot a_n)^{1+\beta} \cdot C_\beta \cdot \left( 1 + C_1 \cdot \sum_{j=1}^{\infty} 2^{j\beta} \exp\left( -C_2 \cdot 4^{j-1} \right) \right) \\
&\leq C_\beta \cdot \left( 1 + \sum_{j=1}^{\infty} 2^{j\beta} \exp\left( -C_2 \cdot 4^{j-1} \right) \right) \cdot (1 + C_1) \cdot (\Lambda(\Phi) \cdot a_n)^{1+\beta}.
\end{aligned}$$

Hence, (3) holds with

$$\tilde{C} = C_\beta \cdot \left( 1 + \sum_{j=1}^{\infty} 2^{j\beta} \exp\left( -C_2 \cdot 4^{j-1} \right) \right) < \infty.$$

This completes the proof of the proposition. ■

**Proof** [Proof of Proposition C.3] The fact that $\mathbb{P}^n \left[A_n(x)\right] \leq \exp(-k/8)$ follows immediately from Lemma B.1 applied with $p = 2k/n$ and $\xi = 1/2$. Moreover, if we take $r_0 = \max\left\{\rho(x, X_i) : i \in S_k^\circ(x, \mathcal{F}_n)\right\}$ then whenever $\mathcal{D}_n \notin A_n(x)$ we have $r_0 \leq r_{2k/n}(x)$, so $\mu\left(B_{r_0}(x)\right) \leq 2k/n$. Hence, letting $r_1 = \max\left\{\rho(x, X_i) : i \in S_k(x, \mathcal{F}_n)\right\}$, since $S$ generates $\omega$ measure-approximate nearest neighbours we have

$$\mu\left(B_{r_1}(x)\right) \leq \omega \cdot \mu\left(B_{r_0}(x)\right) \leq \omega \cdot \frac{2k}{n}.$$

Hence, provided $\mathcal{D}_n \notin A_n(x)$ the fact that $\eta$ is measure-smooth, with constants $(\lambda, C_\lambda)$ implies that for each $i \in S_k(x, \mathcal{F}_n)$ with probability one we have

$$\|\eta(X_i) - \eta(x)\|_\infty \leq C_\lambda \cdot \mu\left(B_{\rho(x, X_i)}(x)\right)^\lambda \leq C_\lambda \cdot \left(\frac{2\omega k}{n}\right)^\lambda.$$

Thus, for $\mathcal{D}_n \notin A_n(x)$,

$$\left\| \frac{1}{k} \sum_{i \in S_k(x, \mathcal{F}_n)} \eta(X_i) - \eta(x) \right\|_\infty \leq C_\lambda \cdot \left(\frac{2\omega k}{n}\right)^\lambda.$$

Hence, for all $\xi \geq 2C_\lambda \cdot \left(\frac{2\omega k}{n}\right)^\lambda$ we have

$$\mathbb{P}^n \left[ \left\| \hat{\eta}^S_{n,k}(x) - \eta(x) \right\|_\infty \geq \xi | \mathcal{D}_n \notin A_n(x) \right]$$
$$\leq \mathbb{P}^n \left[ \left\| \hat{\eta}^S_{n,k}(x) - \frac{1}{k} \sum_{i \in S_k(x,\mathcal{F}_n)} \eta(X_i) \right\|_\infty \geq \frac{\xi}{2} | \mathcal{D}_n \notin A_n(x) \right] \leq 2L \exp(-\frac{k\xi^2}{2}),$$

where the final inequality holds by Lemma B.2, combined with the fact that $\mathbb{P}^n \left[\mathcal{D}_n \notin A_n(x) | \mathcal{F}_n\right] \in \{0,1\}$. ■

## Appendix D. Cost sensitive learning on manifolds

In this section we complete the proof of Theorem 3 by combining Theorems 4 and 5.

**Theorem 3** *Take $d \in \mathbb{N}$ and let $\rho$ denote the Euclidean metric on $\mathbb{R}^d$. Let $\Phi$ be a cost matrix and $\mathcal{M} \subseteq \mathbb{R}^d$ a compact smooth submanifold with dimension $\gamma$ and reach $\tau$. Take positive constants $k_0, r_0, c_0, \nu_{\min}, \nu_{\max}, \zeta_{\max}, \alpha, \beta, C_\alpha, C_\beta$ and let $\Gamma = \langle (c_0, r_0, \nu_{\min}, \nu_{\max}), (\beta, C_\beta, \zeta_{\max}), (\alpha, C_\alpha) \rangle$. Suppose that $S$ generates $\theta$-approximate nearest neighbours for some $\theta \geq 1$. There exists a constant $C > 0$, depending upon $k_0$, $\gamma$, $\tau$, $\Gamma$ such that for all $\mathbb{P} \in \mathcal{P}_\Phi(V_\mathcal{M}, \Gamma)$ and $n \in \mathbb{N}$ the following holds:*

*(1) Given $\xi \in (0,1)$ and $k_n = k_0 \cdot n^{\frac{2\alpha}{2\alpha+\gamma}} \cdot (1 + \log(1/\xi))^{\gamma/(2\alpha+\gamma)}$ with probability at least $1-\xi$ over $\mathcal{D}_n \sim \mathbb{P}^n$ we have*

$$\mathbb{P}\left[f^S_{n,k}(X) \notin \mathcal{Y}^*_\Phi(\eta(X))\right] \leq \xi + C \cdot \left(\theta^\alpha \cdot \Lambda(\Phi) \cdot \sqrt{\log(L)}\right)^\beta \cdot \left(\frac{1+\log(1/\xi)}{n}\right)^{\beta\alpha/(2\alpha+\gamma)}.$$

*(2) Given $k_n = k_0 \cdot n^{\frac{2\alpha}{2\alpha+\gamma}}$ we have*

$$\mathbb{E}_n\left[R\left(f^S_{n,k_n}\right)\right] - R^* \leq C \cdot (\theta^\alpha \cdot \Lambda(\Phi))^{1+\beta} \cdot L \cdot n^{-\frac{\alpha(1+\beta)}{2\alpha+\gamma}}.$$

*Moreover, there exists an absolute constant $K > 0$ such that whenever $\theta > 1$, given any subgaussian random projection $\varphi : \mathbb{R}^d \to \mathbb{R}^h$ with*

$$h \geq K \cdot \|\varphi\|^4_{\psi_2} \cdot \left(\frac{\theta^2+1}{\theta^2-1}\right)^2 \cdot \max\left\{\gamma \log_+(\gamma/(r_0 \cdot \tau)) - \log_+(c_0 \cdot \nu_{\min}) + \gamma, \log \delta^{-1}\right\},$$

*with probability at least $1-\delta$, $S(\varphi)$ generates $\theta$-approximate nearest neighbours, so both (1) and (2) hold with $f^\varphi_{n,k}$ in place of $f^S_{n,k}$.*

We shall require the following lemmas.

**Lemma D.1** *Suppose that $\mathcal{X} = \mathcal{M} \subset \mathbb{R}^d$ is a smooth complete manifold with dimension $\gamma$ and reach $\tau$. Suppose further that $\mathbb{P}$ consists of a marginal $\mu$ which is regular with constants $(c_0, r_0, \nu_{\min}, \nu_{\max})$, along with a conditional label distribution $\eta$ which is Hölder continuous with constants $(\alpha, C_\alpha)$. It follows that the conditional $\eta$ is measure-smooth, with constants $(\lambda, C_\lambda)$, where $\lambda = \alpha/\gamma$ and*

$$C_\lambda = \max\left\{C_\alpha, (\tau/8)^{-\alpha}, (r_0)^{-\alpha}\right\} \cdot \left(c_0 \cdot \nu_{\min} \cdot 4^{-\gamma} \cdot v_\gamma\right)^{-\lambda}.$$

**Proof** We recall from Lemma A.6 that for $x_0 \in \mathcal{M}$ and all $r \leq \tau/8$ we have

$$V_{\mathcal{M}}\left(B_r(x_0)\right) \geq 4^{-\gamma} \cdot v_\gamma \cdot r^\gamma.$$

Moreover, since the marginal $\mu$ which is regular with constants $(c_0, r_0, \nu_{\min}, \nu_{\max})$, for all $r \leq r_0$ we have,

$$\mu\left(B_r(x_0)\right) = \int_{B_r(x_0)} \nu(x)dx \geq \nu_{\min} \cdot V_{\mathcal{M}}\left(\mathrm{supp}(\mu) \cap B_r(x_0)\right) \geq \nu_{\min} \cdot c_0 \cdot V_{\mathcal{M}}\left(B_r(x_0)\right).$$

Thus, for all $r \leq \min\{\tau/8, r_0\}$ we have

$$\mu\left(B_r(x_0)\right) \geq \left(c_0 \cdot \nu_{\min} \cdot 4^{-\gamma} \cdot v_\gamma\right) \cdot r^\gamma.$$

Given $x_0, x_1 \in \mathrm{supp}(\mu)$ we must show that,

$$\|\eta(x_0) - \eta(x_1)\|_\infty \leq C_\lambda \cdot \mu\left(B_{\rho(x_0,x_1)}(x_0)\right)^\lambda.$$

First suppose that $\rho(x_0, x_1) \geq \min\{r_0, \tau/8\}$. Then we have

$$\begin{aligned}\|\eta(x_0) - \eta(x_1)\|_\infty &\leq 1 \leq C_\lambda \cdot \left(\left(c_0 \cdot \nu_{\min} \cdot 4^{-\gamma} \cdot v_\gamma\right) \cdot \left(\min\{\tau/8, r_0\}\right)^\gamma\right)^\lambda \\ &\leq C_\lambda \cdot \mu\left(B_{\min\{r_0,\tau/8\}}(x_0)\right)^\lambda \leq C_\lambda \cdot \mu\left(B_{\rho(x_0,x_1)}(x_0)\right)^\lambda.\end{aligned}$$

One the other hand, if $\rho(x_0, x_1) < \min\{r_0, \tau/8\}$, then

$$\begin{aligned}\|\eta(x_0) - \eta(x_1)\|_\infty &\leq C_\alpha \cdot \rho(x_0, x_1)^\alpha \\ &\leq C_\lambda \cdot \left(c_0 \cdot \nu_{\min} \cdot 4^{-\gamma} \cdot v_\gamma\right)^\lambda \cdot \rho(x_0, x_1)^\alpha \\ &= C_\lambda \cdot \left(\left(c_0 \cdot \nu_{\min} \cdot 4^{-\gamma} \cdot v_\gamma\right) \cdot \rho(x_0, x_1)^\gamma\right)^\lambda \leq C_\lambda \cdot \mu\left(B_{\rho(x_0,x_1)}(x_0)\right)^\lambda.\end{aligned}$$

■

**Lemma D.2** *Suppose that $\mathcal{X} = \mathcal{M} \subset \mathbb{R}^d$ is a smooth complete manifold with dimension $\gamma$ and reach $\tau$. Suppose further that $\mu$ is a regular probability measure with constants $(c_0, r_0, \nu_{\min}, \nu_{\max})$. We let $\tilde{C}$ denote the constant*

$$\tilde{C} = \left(c_0 \cdot \nu_{\min} \cdot 4^{-\gamma} \cdot v_\gamma\right)^{-1} \cdot \max\left\{\nu_{\max} \cdot 4^\gamma \cdot v_\gamma, \left(\min\{\tau/8, r_0\}\right)^{-\gamma}\right\}.$$

*Then, for all $x \in \mathcal{X}$, $r > 0$ and $\theta \geq 1$, we have $\mu\left(B_{\theta \cdot r}(x)\right) \leq \tilde{C} \cdot \theta^\gamma \cdot \mu\left(B_r(x)\right)$.*

**Proof** As noted in the proof of Lemma D.1 given any $x \in \mathrm{supp}(\mu)$ and $r \leq \min\{\tau/8, r_0\}$ we have

$$\mu\left(B_r(x)\right) \geq \left(c_0 \cdot \nu_{\min} \cdot 4^{-\gamma} \cdot v_\gamma\right) \cdot r^\gamma.$$

In addition, by Lemma A.6, for all $x \in \mathcal{X}$ and $r \leq \tau/8$ we have

$$\mu\left(B_r(x)\right) \leq \nu_{\max} \cdot V_{\mathcal{M}}\left(B_r(x)\right) \leq \left(\nu_{\max} \cdot 4^\gamma \cdot v_\gamma\right) \cdot r^\gamma.$$

Now take $x \in \mathcal{X}$, $r > 0$ and $\theta \geq 1$. Firstly, the lemma holds trivially for $x \notin \mathrm{supp}(\mu)$, so we may assume $x \in \mathrm{supp}(\mu)$. We consider two cases.

Case 1: Assume that $\theta \cdot r \leq \min\{\tau/8, r_0\}$, so we have

$$\begin{aligned}\mu\left(B_{\theta \cdot r}(x)\right) &\leq \left(\nu_{\max} \cdot 4^{\gamma} \cdot v_{\gamma}\right) \cdot (\theta \cdot r)^{\gamma} \\ &\leq \left(\left(\nu_{\max} \cdot 4^{\gamma} \cdot v_{\gamma}\right) / \left(c_0 \cdot \nu_{\min} \cdot 4^{-\gamma} \cdot v_{\gamma}\right)\right) \cdot \theta^{\gamma} \cdot \mu\left(B_r(x)\right).\end{aligned}$$

Case 2: Assume that $\theta \cdot r \geq \min\{\tau/8, r_0\}$, so

$$\begin{aligned}\mu\left(B_r(x)\right) &\geq \mu\left(B_{\min\{\tau/8, r_0\}/\theta}(x)\right) \\ &\geq \left(c_0 \cdot \nu_{\min} \cdot 4^{-\gamma} \cdot v_{\gamma}\right) \cdot \min\{\tau/8, r_0\}^{\gamma} \cdot \theta^{-\gamma} \\ &\geq \left(c_0 \cdot \nu_{\min} \cdot 4^{-\gamma} \cdot v_{\gamma}\right) \cdot \min\{\tau/8, r_0\}^{\gamma} \cdot \theta^{-\gamma} \cdot \mu\left(B_{\theta \cdot r}(x)\right).\end{aligned}$$

■

**Lemma D.3** *Suppose that $\mathcal{X} = \mathcal{M} \subset \mathbb{R}^d$ is a smooth complete manifold with dimension $\gamma$ and reach $\tau$. Suppose further that $\mu$ is a regular probability measure with constants $(c_0, r_0, \nu_{\min}, \nu_{\max})$. There exists a constant $\tilde{C}$ which depends purely upon $c_0, r_0, \nu_{\min}, \nu_{\max}$, $\gamma$ and $\tau$ such that given any $\theta \geq 1$, whenever $S$ generates $\theta$ approximate nearest neighbours for some $\theta$. Then $S$ generates $\omega$ measure-approximate nearest neighbours with respect to the measure $\mu$ with $\omega \leq \tilde{C} \cdot \theta^{\gamma}$.*

**Proof** Suppose that $S$ generates $\theta$ approximate nearest neighbours. Take some $n \in \mathbb{N}$, $k \leq n$ and let $r_0 = \max\left\{\rho\left(x, X_i\right) : i \in S_k^{\circ}(x, \mathcal{F}_n)\right\}$ and $r_1 = \max\left\{\rho(x, X_i) : i \in S_k(x, \mathcal{F}_n)\right\}$. Since $S$ generates $\theta$ approximate nearest neighbours we have $r_1 \leq \theta \cdot r_0$. Hence, by Lemma D.2 we have

$$\mu\left(B_{r_1}(x)\right) \leq \mu\left(B_{\theta \cdot r_0}(x)\right) \leq \tilde{C} \cdot \theta^{\gamma} \cdot \mu\left(B_{r_0}(x)\right).$$

■

**Lemma D.4** *Suppose we have a metric space $(\tilde{x}, \tilde{\rho})$ along with a map $\varphi : \mathcal{X} \to \tilde{X}$ together with constants $c_-(\varphi), c_+(\varphi) > 0$ such that for $\mu$ almost every $x_0, x_1 \in \mathcal{X}$ we have*

$$c_-(\varphi) \cdot \rho\left(x_0, x_1\right) \leq \tilde{\rho}\left(\varphi(x_0), \varphi(x_1)\right) \leq c_+(\varphi) \cdot \rho\left(x_0, x_1\right),$$

*where $\tilde{\rho}$ denotes the metric for $\tilde{X}$. For each $k, n \in \mathbb{N}$ we let $S_k(x, \mathcal{F}_n)$ denote the indices of the $k$ nearest neighbours to $\varphi(x)$ in the set $\{\varphi(X_i)\}_{i=1}^n$ with respect to $\tilde{\rho}$. Then $S_k$ generates $\theta$-approximate nearest neighbours with $\theta = c_+(\varphi)/c_-(\varphi)$.*

**Proof** By the construction of $S_k$ we have

$$\begin{aligned}\max\left\{\rho(x, X_i) : i \in S_k(x, \mathcal{F}_n)\right\} &\leq c_-(\varphi)^{-1} \cdot \max\left\{\tilde{\rho}(\varphi(x), \varphi(X_i)) : i \in S_k(x, \mathcal{F}_n)\right\} \\ &\leq c_-(\varphi)^{-1} \cdot \max\left\{\tilde{\rho}(\varphi(x), \varphi(X_i)) : i \in S_k^{\circ}(x, \mathcal{F}_n)\right\} \\ &\leq \left(c_+(\varphi)/c_-(\varphi)\right) \cdot \max\left\{\rho(x, X_i) : i \in S_k^{\circ}(x, \mathcal{F}_n)\right\}.\end{aligned}$$

■

**Proof** [Proof of Theorem 3] As in the statement of Theorem 3 we take a compact smooth submanifold $\mathcal{M} \subseteq \mathbb{R}^d$ with dimension $\gamma$ and reach $\tau$. Take positive constants $k_0, r_0, c_0, \nu_{\min}, \nu_{\max}, \zeta_{\max}, \alpha, \beta, C_\alpha, C_\beta$, and suppose that $S$ generates $\theta$-approximate nearest neighbours for some $\theta \geq 1$. By Lemma D.1 there exists a constant $C_\lambda$, depending upon $\gamma$, $\tau$, $\Gamma$, such that $\eta$ is measure-smooth with constants $(\lambda, C_\lambda)$, where $\lambda = \alpha/\gamma$. In addition, by Lemma D.3 we see that $S$ generates $\omega$-measure approximate nearest neighbours with $\omega \leq \tilde{C} \cdot \theta^\gamma$, where $\tilde{C}$ depends purely upon $\gamma$, $\tau$ and $\Gamma$. Hence, the first part of Theorem 3 follows from Theorem 5.

To prove the second part of Theorem 3, we note that

$$1 = \mu\left(\text{supp}(\mu)\right) = \int_{\text{supp}(\mu)} \nu(x) dV_{\mathcal{M}}(x) \geq \nu_{\min} \cdot V_{\mathcal{M}}\left(\text{supp}(\mu)\right).$$

Hence, $V_{\mathcal{M}}\left(\text{supp}(\mu)\right) \leq \nu_{\min}^{-1}$. Moreover, by assumption supp$(\mu)$ is a $(c_0, r_0)$ regular set. Thus, by Theorem 4, provided

$$h \geq K \cdot \|\varphi\|_{\psi_2}^4 \cdot \left(\frac{\theta^2+1}{\theta^2-1}\right)^2 \cdot \max\left\{\gamma \log_+(\gamma/(r_0\cdot\tau)) - \log_+\left(c_0 \cdot \nu_{\min}\right) + \gamma, \log \delta^{-1}\right\},$$

then with probability at least $1-\delta$, for all pairs $x_0, x_1 \in \text{supp}(\mu)$ we have

$$\frac{2}{\theta^2+1} \cdot \|x_0 - x_1\|_2^2 \leq \|\varphi(x_0) - \varphi(x_1)\|_2^2 \leq \frac{2\theta^2}{\theta^2+1} \cdot \|x_0 - x_1\|_2^2.$$

Hence, with probability at least $1-\delta$, $\varphi : \mathbb{R}^d \to \mathbb{R}^h$ is bi-Lipchitz with constants $c_-(\varphi) = \sqrt{2/(\theta^2+1)}$ and $c_+(\varphi) = \theta \cdot \sqrt{2/(\theta^2+1)}$, so by Lemma D.4, $S(\varphi)$ generates $\theta$ approximate nearest neighbours. ■

## Appendix E. Standard lemmas

Recall that $r_p(x) = \inf\left\{r > 0 : \mu\left(B_r(x)\right) \geq p\right\}$. In this section we prove lemmas B.1 and B.2.

**Lemma B.1** *Take $x \in \mathcal{X}$. Suppose that $p \in [0,1]$, $\xi \leq 1$, $k \leq (1-\xi)np$. Then*

$$\mathbb{P}^n\left[\max\left\{\rho(x, X_i) : i \in S_k^\circ(x, \mathcal{F}_n)\right\} > r_p(x)\right] \leq \exp(-k\xi^2/2).$$

**Proof** Take $r > r_p(x)$, so $\mu\left(B_r(x)\right) \geq p$. Note that $\max\left\{\rho(x, X_i) : i \in S_k^\circ(x, \mathcal{F}_n)\right\} \geq r$ if and only if $\#\mathcal{D}_n \cap B_r(x) < k$, which is equivalent to $\sum_{i=1}^n \mathbb{1}_{\{X_i \in B_r(x)\}} < k$. Moreover, taking

$$\tilde{p} = \frac{1}{n}\sum_{i=1}^n \mathbb{E}\left[\mathbb{1}_{\{X_i \in B_r(x)\}}\right] = \mu\left(B_r(x)\right) \geq p$$

implies $k \leq (1-\xi)n\tilde{p} \leq n\tilde{p}$. Thus, by the multiplicative Chernoff bound (Mitzenmacher and Upfal (2005)) we have,

$$\begin{aligned}\mathbb{P}^n\left[\rho_\infty\left(x, \{X_i : i \in S_k^\circ(x, \mathcal{F}_n)\}\right) \geq r\right] &\leq \mathbb{P}\left[\sum_{i=1}^n \mathbb{1}_{\{X_i \in B_r(x)\}} < k\right] \\ &\leq \mathbb{P}\left[\sum_{i=1}^n \mathbb{1}_{\{X_i \in B_r(x)\}} < (1-\xi)n\tilde{p}\right] \\ &\leq \exp(-n\tilde{p}\xi^2/2) \leq \exp(-k\xi^2/2).\end{aligned}$$

Since this holds for any countable decreasing sequence of $r$ with each $r > r_p(x)$, the lemma follows by continuity. ■

**Lemma B.2** *Take $x \in \mathcal{X}$. For each $\delta > 0$ we have*

$$\mathbb{P}^n \left[ \left\| \hat{\eta}^S_{n,k}(x) - \frac{1}{k} \sum_{i \in S_k(x, \mathcal{F}_n)} \eta(X_i) \right\|_\infty \geq \delta | \mathcal{F}_n \right] \leq 2L \exp(-2k\delta^2).$$

**Proof** By the construction of $\hat{\eta}^S_{n,k}(x)$ it suffices to show that

$$\mathbb{P}^n \left[ \left\| \sum_{i \in S_k(x, \mathcal{F}_n)} (e(Y_i) - \eta(X_i)) \right\|_\infty \geq k \cdot \delta | \mathcal{F}_n \right] \leq 2L \exp(-2k\delta^2).$$

By the union bound it suffices to show that for each $l \in \mathcal{Y}$ we have

$$\mathbb{P}^n \left[ \left| \sum_{i \in S_k(x, \mathcal{F}_n)} (e(Y_i)_l - \eta(X_i)_l) \right| \geq k \cdot \delta | \mathcal{F}_n \right] \leq 2 \exp(-2k\delta^2).$$

This in turn follows if we show that for all $x_1, \cdots, x_n \in \mathcal{X}$ we have

$$\mathbb{P}^n \left[ \left| \sum_{i \in S_k(x, \mathcal{F}_n)} (e(Y_i)_l - \eta(x_i)_l) \right| \geq k \cdot \delta \middle| X_1 = x_1, \cdots, X_n = x_n \right] \leq 2 \exp(-2k\delta^2).$$

Moreover, $e(Y_i)_l = \mathbb{1}_{\{Y_i = l\}}$ and $\eta(x_i)_l = \mathbb{P}[Y_i = l | X_i = x_i] = \mathbb{E}\left[\mathbb{1}_{\{Y_i = l\}} | X_i = x_i\right]$. Thus, we must show that

$$\mathbb{P}^n \left[ \left| \sum_{i \in S_k(x, \mathcal{F}_n)} \left(\mathbb{1}_{\{Y_i = l\}} - \mathbb{E}\left[\mathbb{1}_{\{Y_i = l\}} | X_i = x_i\right]\right) \right| \geq k \cdot \delta \middle| X_1 = x_1, \cdots, X_n = x_n \right]$$

does not exceed $2 \exp(-2k\delta^2)$. This is immediate from Hoeffding's inequality (Boucheron et al. (2013)). ■

## Appendix F. Geometric lemmas

**Lemma A.6** *Let $\mathcal{M} \subseteq \mathbb{R}^d$ be a compact smooth submanifold with dimension $\gamma$, reach $\tau$ and Riemannian volume form $V_{\mathcal{M}}$. Then for all $x \in \mathcal{M}$ and $r < \tau/8$ we have*

$$4^{-\gamma} \cdot v_\gamma \cdot r^\gamma \leq V_{\mathcal{M}}\left(B^g_r(x)\right) \leq V_{\mathcal{M}}\left(B_r(x)\right) \leq 4^\gamma \cdot v_\gamma \cdot r^\gamma.$$

**Proof** Fix $x \in \mathcal{M}$ and $r < \tau/8$. By (Chazal, 2013, Corollary 1.3) we have

$$V_{\mathcal{M}}\left(B_r(x)\right) \leq \left(\frac{2\tau}{\tau - 4r}\right)^\gamma \cdot v_\gamma \cdot r^\gamma \leq 4^\gamma \cdot v_\gamma \cdot r^\gamma.$$

In addition by (Eftekhari and Wakin, 2015, Lemma 12) we have

$$V_{\mathcal{M}}\left(B_r(x)\right) \geq \left(1-\frac{r^2}{4\tau^2}\right)^{\frac{\gamma}{2}} \cdot v_\gamma \cdot r^\gamma \geq v_\gamma \cdot 2^{-\gamma} \cdot r^\gamma. \tag{4}$$

Now suppose $z \in B_{r/2}(x)$, so $\|z-x\| < r/2$, then by (Niyogi et al., 2008, Proposition 6.3) we have

$$\rho_g(z,x) \leq \tau - \tau\sqrt{1-\frac{2\|z-x\|_2}{\tau}} \leq 2 \cdot \|z-x\|_2 < r.$$

Hence, $B_{r/2}(x) \subseteq B^g_r(x)$. Hence, by (4) we have

$$V_{\mathcal{M}}\left(B^g_r(x)\right) \geq V_{\mathcal{M}}\left(B_{r/2}(x)\right) \geq v_\gamma \cdot 4^{-\gamma} \cdot r^\gamma.$$

■

**Lemma A.7** *With the assumptions of lemma A.6, for all $x, \tilde{x} \in \mathcal{M}$ and $\tilde{r} \leq r < \tau/8$ with $\rho_g(x,\tilde{x}) \leq r + \tilde{r}/2$ we have $V_{\mathcal{M}}\left(B^g_r(x) \cap B^g_{\tilde{r}}(\tilde{x})\right) \geq 2^{-4\gamma} \cdot v_\gamma \cdot \tilde{r}^\gamma$.*

**Proof** Fix $x \in \mathcal{M}$ and $r > 0$ and take $\tilde{x} \in \overline{B^g_r(x)}$ and $\tilde{r} \in (0, \min\{r, \tau/8\})$. We claim that there exists a geodesic ball $B^g_{\tilde{r}/4}(y)$ of radius $\tilde{r}/2$ such that $B^g_{\tilde{r}/2}(y) \subseteq B^g_r(x) \cap B^g_{\tilde{r}}(\tilde{x})$. To see this consider two cases.

Case 1: Suppose that $\rho_g(x,\tilde{x}) \leq 3\tilde{r}/4 \leq 3r/4$. Then given $z \in B^g_{\tilde{r}/4}(\tilde{x})$ we have $\rho_g(x,z) \leq \rho_g(x,\tilde{x}) + \rho_g(\tilde{x},z) < 3r/4 + \tilde{r}/4 \leq r$. Hence, $B^g_{\tilde{r}/4}(\tilde{x}) \subseteq B^g_r(x) \cap B^g_{\tilde{r}}(\tilde{x})$, so the claim holds with $y = \tilde{x}$.

Case 2: Suppose that $3\tilde{r}/4 < \rho_g(x,\tilde{x}) \leq r$. Let $c : [0, \rho_g(x,\tilde{x})] \to \mathcal{M}$ be a unit speed geodesic with $c(0) = x$ and $c(1) = \tilde{x}$. Let $y = c\left(\rho_g(x,\tilde{x}) - 3\tilde{r}/4\right)$, so $\rho_g(y,\tilde{x}) = 3\tilde{r}/4$ and $\rho_g(y,x) = \rho_g(x,\tilde{x}) - 3\tilde{r}/4 \leq r - \tilde{r}/4$. Hence, $B^g_{\tilde{r}/4}(y) \subseteq B^g_r(x) \cap B^g_{\tilde{r}}(\tilde{x})$.

Hence, the claim holds. Thus, by two applications of Lemma A.6 we have,

$$V_{\mathcal{M}}\left(B^g_r(x) \cap B^g_{\tilde{r}}(\tilde{x})\right) \geq V_{\mathcal{M}}\left(B^g_{\tilde{r}/4}(y)\right) \geq 2^{-4\gamma} \cdot v_\gamma \cdot \tilde{r}^\gamma.$$

■

## Appendix G. Random projections theorem

Our goal in this section is to prove Theorem 4.

**Theorem 4** *There exists an absolute constant $K$ such that the following holds. Given a compact smooth submanifold $\mathcal{M} \subseteq \mathbb{R}^d$ with dimension $\gamma$ and reach $\tau$, suppose that $A \subset \mathcal{M}$ is $(c_0, r_0)$ regular with respect to the Riemannian volume $V_{\mathcal{M}}$. Suppose that $\varphi : \mathbb{R}^d \to \mathbb{R}^h$ is a subgaussian random projection. Take $\epsilon, \delta \in (0,1)$ and suppose that*

$$h \geq K \cdot \|\varphi\|^4_{\psi_2} \cdot \epsilon^{-2} \cdot \max\left\{\gamma \log_+(\gamma/(r_0 \cdot \tau)) + \log_+\left(V_{\mathcal{M}}(A)/c_0\right) + \gamma, \log \delta^{-1}\right\}.$$

*Then with probability at least $1-\delta$, for all pairs $x_0, x_1 \in A$ we have*

$$(1-\epsilon) \cdot \|x_0 - x_1\|^2_2 \leq \|\varphi(x_0) - \varphi(x_1)\|^2_2 \leq (1+\epsilon) \cdot \|x_0 - x_1\|^2_2.$$

The result generalises Theorem 7.9 from Dirksen (2016) and the proof is very similar. We begin by recalling another important result from Dirksen (2016). We require some notation. Given a metric space $(\mathcal{X}, \rho)$ and a subset $A \subset \mathcal{X}$ we define

$$\gamma_{\text{tal}}(A) := \inf_{\mathcal{T}} \left\{ \sup_{x \in A} \left\{ \sum_{q \geq 0} 2^{q/2} \cdot \rho(x, T_q) \right\} \right\},$$

where the infimum is taken over sequences $\mathcal{T} = \{T_q\}_{q \in \mathbb{N}}$ with each $T_q \subset A$, $\#(T_0) = 1$ and for $q \geq 1$, $\#(T_q) \leq 2^{2^q}$. Given $x_0, x_1 \in \mathbb{R}^d$ we let $\text{Ch}(x_0, x_1)$ denote the normalised chord,

$$\text{Ch}(x_0, x_1) = \frac{x_0 - x_1}{\|x_0 - x_1\|_2}.$$

Given a set $A \subset \mathbb{R}^d$ we let $A_{\text{nc}} \subset \mathbb{S}^{d-1} \subset \mathbb{R}^d$ denote the set of normalised chords, $A_{\text{nc}} = \{\text{Ch}(x_0, x_1) : x_0, x_1 \in A\}$. Given $x \in \mathcal{M}$ we let $P_x$ denote the projection onto the tangent space of $\mathcal{M}$ at $x$. Given a matrix $M$ we let $\|M\|_{\text{op}}$ denote the operator norm of $M$. Given a semi-metric space $(\mathcal{X}, \rho)$, a subset $A \subset \mathcal{X}$ and $r > 0$, a subset $\{x_1, \cdots, x_q\} \subset A$ such that for each $a \in A$ we have $\rho(a, x_i) < r$ for some $i \in \{1, \cdots, q\}$ is referred to as an $r$-net of $A$ with respect to $\rho$. We let $N(A, \rho, r)$ denote the cardinality of the smallest $r$-net of $A$ with respect to $\rho$.

**Theorem 6 (Dirksen (2016))** *There exists an absolute constant $K$ such that the following holds. Suppose that $A \subset \mathbb{R}^d$ and let $\varphi : \mathbb{R}^d \to \mathbb{R}^h$. Take $\epsilon, \delta \in (0,1)$ and suppose that*

$$h \geq K \cdot \|\varphi\|_{\psi_2}^4 \cdot \epsilon^{-2} \cdot \max\left\{\gamma_{tal}^2(A_{nc}), \log\left(\delta^{-1}\right)\right\}.$$

*Then with probability at least $1 - \delta$, for all pairs $x_0, x_1 \in A$ we have*

$$(1 - \epsilon) \cdot \|x_0 - x_1\|_2^2 \leq \|\varphi(x_0) - \varphi(x_1)\|_2^2 \leq (1 + \epsilon) \cdot \|x_0 - x_1\|_2^2.$$

Given a $(c_0, r_0)$ regular set $A \subset \mathcal{M}$ we shall seek to bound $\gamma_{\text{tal}}^2(A_{\text{nc}})$. We shall use the following upper bound.

**Lemma G.1 (Talagrand (2006))** *Given a metric space $(\mathcal{X}, \rho)$ and a subset $A \subset \mathcal{X}$ we have*

$$\gamma_{tal}(A) \leq (\log 2)^{-1/2} \cdot \int_0^{diam(A)} \sqrt{\log N(A, \rho, r)} dr.$$

**Proof** See (Talagrand, 2006, pg. 13). ■

We require the following lemmas from Dirksen (2016).

**Lemma G.2 (Dirksen (2016))** *Suppose that $\mathcal{M} \subset \mathbb{R}^d$ is a manifold with reach $\tau$. Then, given any $x_0, x_1, y_0, y_1 \in \mathcal{M}$ we have,*

*(a)* $\|Ch(x_0, x_1) - Ch(y_0, y_1)\|_2 \leq 2 \cdot (\|x_0 - y_0\|_2 + \|x_1 - y_1\|_2) / (\|x_0 - x_1\|_2)$,

*(b)* $\|Ch(x_0, x_1) - P_{x_0}(Ch(x_0, x_1))\|_2 \leq 2\tau^{-1} \cdot \|x_0 - x_1\|_2$,

*(c)* $\|P_{x_0} - P_{x_1}\|_{op} \leq 2\sqrt{2} \cdot \tau^{-1/2} \cdot \|x_0 - x_1\|_2^{1/2}$.

**Lemma G.3** *Take any subset $A \subset \mathcal{M}$ where $\mathcal{M} \subset \mathbb{R}^d$ is a $\gamma$-dimensional manifold of reach $\tau$. For all $r > 0$, if we let $n(r) = N\left(A, \|\cdot\|_2, \min\{\tau, 1/8\} \cdot r^2/16\right)$ then we have,*

$$N\left(A_{nc}, \|\cdot\|_2, r\right) \leq n(r) \cdot \left(n(r) + \left(1 + \frac{4}{r}\right)^{\gamma}\right).$$

**Proof** Take $a, b > 0$ and let $\{x_1, \cdots, x_q\} \subset A$ be a minimal $a$-net of $A$ with respect to $\|\cdot\|_2$, and for each $i = 1, \cdots, q$ we let $\{y_{ij}\}_{j=1}^{m_i}$ be a $b$-net for $A$ with respect to the semi-metric $(z_0, z_1) \mapsto \|P_{x_i}(z_0 - z_1)\|_2$. Note that $q = N\left(A, \|\cdot\|_2, a\right)$ and for each $i$,

$$m_i = N\left(A, \|P_{x_i}\left(\cdot\right)\|_2, b\right) \leq N\left(\{w \in \mathbb{R}^{\gamma} : \|w\|_2 \leq 1\}, \|\cdot\|_2, b\right) \leq \left(1 + \frac{2}{b}\right)^{\gamma}.$$

Now take $t > 0$ and decompose $A_{\text{nc}}$ into $A_{\text{nc}}^{\geq t}$ and $A_{\text{nc}}^{<t}$ by

$$A_{\text{nc}}^{\geq t} = \{\text{Ch}(z_0, z_1) : z_0, z_1 \in A, \|z_0 - z_1\| \geq t\}$$
$$A_{\text{nc}}^{<t} = \{\text{Ch}(z_0, z_1) : z_0, z_1 \in A, \|z_0 - z_1\| < t\}.$$

Given $z_0, z_1 \in A$ with $\|z_0 - z_1\| \geq t$ we may take $i_0, i_1 \in \{1, \cdots, q\}$ so that $\|z_0 - x_{i_0}\|_2 < a$ and $\|z_1 - x_{i_1}\|_2 < a$. It follows from Lemma G.2 (a) that $\|\text{Ch}(z_0, z_1) - \text{Ch}(x_{i_0}, x_{i_1})\|_2 < 2a/t$. Thus, we have

$$N\left(A_{\text{nc}}^{\geq t}, \|\cdot\|_2, 2a/t\right) \leq N\left(A, \|\cdot\|_2, a\right)^2.$$

Given $z_0, z_1 \in A$ with $\|z_0 - z_1\| < t$ we may take $i_0 \in \{1, \cdots, q\}$ so that $\|z_0 - x_{i_0}\|_2 < a$ and $j \in \{1, \cdots, m_{i_0}\}$ so that $\|P_{x_{i_0}}\left(\text{Ch}(z_0, z_1) - y_{i_0 j}\right)\|_2 < b$. Hence,

$$\begin{aligned}\|\text{Ch}(z_0, z_1) - P_{x_{i_0}}\left(y_{i_0 j}\right)\|_2 \leq\ & \|\text{Ch}(z_0, z_1) - P_{z_0}\left(\text{Ch}(z_0, z_1)\right)\|_2 \\ & + \|P_{z_0}\left(\text{Ch}(z_0, z_1)\right) - P_{x_{i_0}}\left(\text{Ch}(z_0, z_1)\right)\|_2 \\ & + \|P_{x_{i_0}}\left(\text{Ch}(z_0, z_1) - y_{i_0 j}\right)\|_2 \\ & \leq 2\tau^{-1} \cdot t + 2\sqrt{2} \cdot a^{1/2} + b,\end{aligned}$$

by applying Lemma G.2 (b) and (c). Thus, we have,

$$\begin{aligned}N\left(A_{\text{nc}}^{<t}, \|\cdot\|_2, \left(2\tau^{-1} \cdot t + 2\sqrt{2} \cdot a^{1/2} + b\right)\right) & \leq N\left(A, \|\cdot\|_2, a\right) \cdot \max_i\{N\left(A, \|P_{x_i}\left(\cdot\right)\|_2, b\right)\} \\ & \leq N\left(A, \|\cdot\|_2, a\right) \cdot \left(1 + \frac{2}{b}\right)^{\gamma}.\end{aligned}$$

Hence, given any $r > 0$, taking $a = \min\{\tau, 1/8\} \cdot r^2/16$, $b = r/2$ and $t = 2a/r$ and letting $n(r) = N\left(A, \|\cdot\|_2, \min\{\tau, 1/8\} \cdot r^2/16\right)$, we have

$$N\left(A_{\text{nc}}, \|\cdot\|_2, r\right) \leq n(r) \cdot \left(n(r) + \left(1 + \frac{4}{r}\right)^{\gamma}\right).$$

■

**Lemma G.4** *Take any subset $A \subset \mathcal{M}$ where $\mathcal{M} \subset \mathbb{R}^d$ is a $\gamma$-dimensional manifold of reach $\tau$. Suppose further that $A$ is a $(c_0, r_0)$-regular set. Then, for all $r < \min\{r_0, \tau/2\}$ we have*

$$N\left(A, \|\cdot\|_2, r\right) \leq c_0^{-1} \cdot V_{\mathcal{M}}\left(A\right) \cdot (\gamma+4)^{\gamma/2+2} \cdot r^{-\gamma}.$$

**Proof** Let $\{x_1, \cdots, x_q\} \subset A$ be a maximal $r$-separated set. By (Eftekhari and Wakin, 2015, Lemma 12), for each $i$ we have

$$\begin{aligned} V_{\mathcal{M}}\left(B_{r/2}(x_i)\right) &\geq (63/64)^{\gamma/2} \cdot v_\gamma \cdot (r/2)^\gamma = \left(((63\cdot\pi)/256)^{\gamma/2}/\Gamma\left(\frac{\gamma}{2}+1\right)\right) \cdot r^\gamma \\ &\geq \left(((63\cdot\pi)/256)^{\gamma/2} \cdot \left(\frac{\gamma}{2}+2\right)^{-(\gamma/2+2)}\right) \cdot r^\gamma \geq (\gamma+4)^{-(\gamma/2+2)} \cdot r^\gamma. \end{aligned}$$

Since $\{x_1, \cdots, x_q\} \subset A$ is $r$-separated, the ballse $B_{r/2}(x_i)$ are disjoint, so using the $(c_0, r_0)$ regularity property,

$$V_{\mathcal{M}}\left(A\right) \geq \sum_{i=1}^q V_{\mathcal{M}}\left(A \cap B_{r/2}(x_i)\right) \geq c_0 \cdot \sum_{i=1}^q V_{\mathcal{M}}\left(B_{r/2}(x_i)\right) \geq q \cdot c_0 \cdot (\gamma+4)^{-(\gamma/2+2)} \cdot r^\gamma.$$

Moreover, since $\{x_1, \cdots, x_q\} \subset A$ is a maximal $r$-separated set, $\{x_1, \cdots, x_q\}$ must also be an $r$-net for $A$ with respect to $\|\cdot\|_2$, so $N\left(A, \|\cdot\|_2, r\right) \leq q$. Thus,

$$N\left(A, \|\cdot\|_2, r\right) \leq c_0^{-1} \cdot V_{\mathcal{M}}\left(A\right) \cdot (\gamma+4)^{\gamma/2+2} \cdot r^{-\gamma}.$$

■

**Lemma G.5** *There exists a universal constant $\tilde{K} > 0$ such that the following holds. Take any subset $A \subset \mathcal{M}$ where $\mathcal{M} \subset \mathbb{R}^d$ is a $\gamma$-dimensional manifold of reach $\tau$. Suppose further that $A$ is a $(c_0, r_0)$-regular set. Then, for all $r < \{r_0, 1\}$ we have*

$$\log\left(N\left(A_{nc}, \|\cdot\|_2, r\right)\right) \leq \tilde{K} \cdot \left(\gamma \log_+(\gamma/\tau) + \log_+\left(V_{\mathcal{M}}(A)/c_0\right) - \gamma \log_+(r) + \gamma\right).$$

**Proof** Combine lemmas G.3 and G.4. ■

**Proof** [Proof of Theorem 4] To complete the proof of Theorem 4, we apply Lemmas G.1 and G.5,

$$\begin{aligned} \left(\gamma_{\text{tal}}\left(A_{\text{nc}}\right)\right)^2 &\leq (\log 2)^{-1} \cdot \left(\int_0^2 \sqrt{\log N(A_{\text{nc}}, \rho, r)}dr\right)^2 \\ &\leq 2 \cdot (\log 2)^{-1} \cdot \int_0^2 \log N(A_{\text{nc}}, \rho, r)dr \\ &\leq K' \cdot \left(\gamma \log_+(\gamma/(\tau r_0)) + \log_+\left(V_{\mathcal{M}}(A)/c_0\right) + \gamma\right). \end{aligned}$$

Hence, Theorem 4 follows from Theorem 6. ■

## Appendix H. Notation

| | |
|---|---|
| $\zeta_{\max}$ | Upper limit for $\zeta$ in the margin condition |
| $r_0$ | Regularity radius for $\mathrm{supp}(\mu)$ |
| $c_0$ | Regularity coefficient $\mathrm{supp}(\mu)$ |
| $\mathcal{Y}_\Phi^*$ | The set of labels which minimise the cost sensitive loss for a given conditional distribution |
| $f_{n,k}^S$ | A classifier based on $n$ training examples and $k$ approximate nearest neighbours generated via $S_k$ |
| $\hat{\eta}_{n,k}^S$ | An estimator of $\eta$ based on $n$ training examples and $k$ approximate nearest neighbours generated via $S_k$ |
| $k$ | The number of nearest neighbours |
| $\mathcal{X}$ | The feature space |
| $Z_i$ | The $i$th training example from $\mathcal{D}_n$ |
| $X_i$ | The $i$th feature vector in the training data |
| $Y_i$ | The $i$th test label in the training data |
| $\mathcal{Y}$ | The set of class labels |
| $\mathcal{Z}$ | The Cartesian product $\mathcal{X} \times \mathcal{Y}$ |
| $\mathcal{D}_n$ | A data set $\mathcal{D}_n = \{Z_1, \cdots, Z_n\}$ of size $n$, with each $Z_i = (X_i, Y_i) \sim \mathbb{P}$ chosen independently |
| $n$ | Number of training examples |
| $S_k^\circ$ | A function from a point to the indices of its $k$ nearest neighbours |
| $S_k$ | A function from a point indices of a set of approximate $k$ nearest neighbours |
| $\theta$ | Scale factor for approximate nearest neighbours |
| $\omega$ | Scale factor for measure-approximate nearest neighbours |
| $\lambda$ | Smoothness exponent for $\eta$ with respect to $\mu$ |
| $\alpha$ | Hölder exponent for $\eta$ |
| $C_\alpha$ | Hölder scaling constant for $\eta$ |
| $\gamma$ | Dimension of the manifold $\mathcal{M}$ |
| $\Gamma$ | Set of parameters defining a set of distributions on $\mathcal{M}$ |
| $V_{\mathcal{M}}$ | The Riemannian volume form on manifold $\mathcal{M}$ |
| $\nu$ | The density of $\mu$ with respect $V_{\mathcal{M}}$ |
| $\beta$ | Margin exponent for $\mathbb{P}$ |
| $C_\beta$ | Margin scaling constant for $\mathbb{P}$ |
| $C_\lambda$ | Smoothness scaling constant for $\eta$ |
| $x$ | A feature vector |
| $y$ | A class label |
| $X$ | A random feature vector |
| $Y$ | A random class label |
| $e(y)$ | A $L \times 1$ one-hot-encoding of the class label $y$ |
| $\mathcal{F}_n$ | The ordered set $\{X_1, \cdots, X_n\}$ where $\mathcal{D}_n = \{(X_1, Y_1), \cdots, (X_n, Y_n)\}$ |
| $\mathbb{P}$ | Distribution over $\mathcal{Z} = \mathcal{X} \times \mathcal{Y}$ |
| $\mu$ | The marginal distribution over $\mathcal{X}$ ie. $\mu(A) = \mathbb{P}[X \in A]$ for $A \subseteq \mathcal{X}$ |
| $\eta$ | The conditional distribution of $Y$ given $X = x$, as a probability vector |
| $L$ | Number of classes |
| $\Phi$ | A $L \times L$ cost matrix with entries $\phi_{i,j}$ |
| $\phi_{i,j}$ | The cost incurred by predicting class $i$ when the true label is class $j$ |
| $M_\Phi$ | The cost-sensitive margin |
| $\mathbb{P}^n$ | Probability over data sets $\mathcal{D}_n$ of size $n$ with each $(x_i, y_i)$ sampled i.i.d from $\mathbb{P}$ |
| $\mathbb{E}_n$ | Expectation over data sets $\mathcal{D}_n$ according to $\mathbb{P}^n$ |
| $\rho$ | A metric on $\mathcal{X}$. If $\mathcal{X} \subset \mathbb{R}^d$ then $\rho$ denotes the Euclidean metric. |
| $\rho_g$ | The Riemannian metric on $\mathcal{M}$ |
| $\|\cdot\|_2$ | The Euclidean norm on $\mathbb{R}^d$ |
| $R(h)$ | The risk of a classifier $h$ |
| $R^*$ | The Bayes risk |
| $B_r(x)$ | The open metric ball of radius $r$ centered at $x$ |
| $\overline{B_r(x)}$ | The closed metric ball of radius $r$ centered at $x$ |
| $B_r^g(x)$ | The open metric ball of radius $r$ centered at $x$ with respect to the geodesic metric $\rho_g$ |
| $\overline{B_r^g(x)}$ | The closed metric ball of radius $r$ centered at $x$ with respect to the geodesic metric $\rho_g$ |
| $\mathbb{1}_A$ | The indicator function of a set $A$ |
| $\tau$ | Reach for a Riemannian manifold $\mathcal{M}$ |
| $\mathcal{M}$ | A compact $C^\infty$-smooth submanifold of $\mathbb{R}^d$ |
| $d$ | Dimension of the ambient Euclidean space $\mathbb{R}^d$ |
| $\mathbb{R}^d$ | $d$ dimensional Euclidean space |
| Asym | The asymmetry of a cost matrix |
| $\mathcal{P}_\Phi(\upsilon, \Gamma)$ | A class of measures specified in Definition 2.5 |
| $\Lambda(\Phi)$ | The constant $\Lambda(\Phi) := (L-2) \cdot \text{Asym}(\Phi) + 2 \cdot \lVert\Phi\rVert_\infty$ |